\documentclass[sigconf]{acmart}

\usepackage{tikz}
\usepackage{amsmath}
\usepackage[skip=2pt]{caption}   
\usepackage{hyperref}
\usepackage{algorithm,algorithmic}
\usepackage{graphicx}
\usepackage{setspace}
\usepackage{textcomp}
\usepackage{xcolor}
\usepackage{soul}
\usepackage{listings}
\usepackage{tikz}
\usepackage{pgfplots}
\usepackage{dblfloatfix}
\usepackage{float}
\usepackage{pgfplots}
\usepackage{tikz}
\pgfplotsset{compat=1.18}
\usetikzlibrary{shapes.geometric, arrows.meta, positioning, calc, shadows, fit, backgrounds}

\definecolor{mutedgreen}{RGB}{33, 121, 33}

\definecolor{stepblue}{RGB}{235, 242, 250}
\definecolor{stepgray}{RGB}{245, 245, 245}
\definecolor{stepdark}{RGB}{60, 60, 70}
\definecolor{stepaccent}{RGB}{0, 85, 160}
\definecolor{loopbg}{RGB}{250, 250, 252}
\definecolor{stepinit}{RGB}{225, 255, 240} 

\title{Stochastic Event Prediction via Temporal Motif Transitions}

\author{İbrahim Bahadır Altun}
\affiliation{
  \institution{University at Buffalo}
  \country{}}
\email{ialtun@buffalo.edu}

\author{Ahmet Erdem Sarıyüce}
\affiliation{
  \institution{University at Buffalo}
  \country{}}
\email{erdem@buffalo.edu}


\setcopyright{none}
\AtBeginDocument{\renewcommand{\footnotetextcopyrightpermission}[1]{}}

\begin{document}

\begin{abstract}
Networks of timestamped interactions arise across social, financial, and biological domains, where forecasting future events requires modeling both evolving topology and temporal ordering.
Temporal link prediction methods typically frame the task as binary classification with negative sampling, discarding the sequential and correlated nature of real-world interactions. We introduce STEP (STochastic Event Predictor), a framework that reformulates temporal link prediction as a sequential forecasting problem in continuous time. STEP models event dynamics through discrete temporal motif transitions governed by Poisson processes, maintaining a set of open motif instances that evolve as new interactions arrive. At each step, the framework decides whether to initiate a new temporal motif or extend an existing one, selecting the most probable event via Bayesian scoring of temporal likelihoods and structural priors. STEP also produces compact, temporal motif-based feature vectors that can be concatenated with existing temporal graph neural network outputs, enriching their representations without architectural modifications. Experiments on five real-world datasets demonstrate up to 21\% average precision gains over state-of-the-art baselines in classification and 0.99 precision in next $k$ sequential forecasting, with consistently lower runtime than competing motif-aware methods.
%
\end{abstract}

\maketitle
\fancyhead{}
\fancyfoot{}

\section{Introduction}
Networks of timestamped interactions underpin social media, financial systems, cybersecurity telemetry, and biological pathways. Each interaction is an event occurring at a precise time, so effective forecasting must respect both evolving topology and temporal ordering. A particular challenge in this area is \emph{Temporal Link Prediction}, which seeks to infer future events based on historical graph dynamics. Traditional approaches focus on learning fixed embeddings for events or nodes and then applying a classifier to predict whether a given link will appear. While effective for static or mildly evolving graphs, these methods do not directly model how future interactions are generated in continuous time.

Early link prediction methods aggregate events into snapshots and apply structural heuristics such as common neighbors, Adamic-Adar, preferential attachment, or Katz \cite{libenowell2003,adamic2003,barabasi1999,katz1953}. Modern static approaches learn embeddings via matrix factorization or random walks \cite{menon2011,perozzi2014,grover2016}, but time aggregation discards fine-grained ordering.
More recently, continuous-time models based on Poisson or Hawkes processes preserve event ordering \cite{hawkes1971} yet struggle to scale to large-scale graphs or to incorporate expressive structural features. Temporal graph neural networks (TGNNs) learn message-passing dynamics over event streams (e.g., DyRep, TGAT, TGN, EvolveGCN) \cite{trivedi2019dyrep,xu2020tgat,rossi2020tgn,pareja2020evolvegcn} but remain tied to classification-centric objectives and often require exhaustive candidate evaluation. Motif-aware representations improve temporal expressiveness \cite{zhao2022motif}, yet most methods still optimize binary classification with negative sampling, limiting their utility for time-aware, ordered forecasts.

Beyond modeling limitations, the standard evaluation paradigm itself introduces significant concerns. Negative sampling strategies that draw from all possible node pairs inflate metrics by including trivial negatives unlikely to arise in practice \cite{Poursafaei2022}, while the assumption that future edges are independent binary decisions disregards the sequential and correlated nature of real-world interactions \cite{Lampert2024}. Efforts such as the Temporal Graph Benchmark \cite{Huang2023} have improved evaluation through ranking-based metrics, but still rely on predefined candidate sets. Lampert et al. \cite{Lampert2024} further show that batch-based pipelines induce artificial temporal semantics, revealing that many reported gains stem from evaluation artifacts rather than genuine predictive capability. However, their work primarily serves as a diagnostic framework. 

To address these challenges, we reformulate temporal link prediction as a sequential forecasting problem, better reflecting real-world scenarios such as notification scheduling, event prioritization, and network simulation \cite{Du2016RMTPP,Kumar2019,zhou2020taggen} where models must reason under uncertainty and respect event ordering. We present STEP (STochastic Event Predictor), a framework that models temporal graph evolution through discrete temporal motif transitions governed by Poisson processes. At each step, STEP decides whether to initiate a new temporal motif or extend an existing one, selecting the most probable event via Bayesian scoring of temporal likelihoods and structural priors. STEP also produces compact, temporal motif-based feature vectors that can be concatenated with TGNN outputs to enrich their representations without architectural modifications, while still supporting standard classification evaluation.
%
%
Our key contributions are as follows:



\begin{itemize}
\item We formulate temporal link prediction as a forecasting problem rather than pure binary classification, which sidesteps the need for exhaustive negative sampling while still allowing comparison with standard classification benchmarks.
\item We propose a lightweight generative model that combines Poisson-driven event arrivals with Bayesian scoring over motif transitions, enabling sequential next $k$ prediction without neural network training.
\item We introduce compact, temporal motif-based probabilistic feature vectors that can be combined with TGNN prediction scores, enriching event representations without requiring any modifications to existing TGNN architectures.
\item We conduct extensive experiments on five real-world datasets, demonstrating up to 21\% average precision gains over TGNN baselines in classification and 0.99 precision in next $k$ sequential forecasting.
\item We provide an efficient C++ implementation that interfaces with Python-based TGNNs via inter-process communication, enabling low-latency inference on graphs with up to seven million events.
\end{itemize}

\section{Related Work}

Temporal link prediction methods differ in how they model event dynamics and how they capture candidate interactions. Despite architectural differences, most temporal link prediction methods consider binary classification for evaluation. Typically, these methods aim to predict whether an edge between two nodes will occur within a specified future time. They use chronological data splits for training, validation, and testing, and rely on negative sampling to balance the dataset. In this section, we provide a concise review of recent modeling trends, emphasizing their architectural choices and potential drawbacks. 

\noindent \textbf{Point-process models. }
Point-processes treat interaction histories as continuous event streams, where the probability of a future event is determined by an intensity function. Hawkes processes define event intensities that rise or decay over time, yielding calibrated probabilities \cite{hawkes1971,asmussen2007} but require intensities for many node pairs. Neural variants approximate intensities \cite{mei2017neural} and spatio-temporal processes incorporate context \cite{zarezade2017}, but struggle to encode higher-order patterns without significant computational cost. Recent models introduce structure-informed excitation functions \cite{Zuo2020}, yet remain bottlenecked by high per-step computation. In contrast, STEP overcomes these bottlenecks by modeling events through discrete motif transitions rather than pair-wise intensities. This enables capturing higher-order structural patterns with minimal runtime overhead.

\noindent \textbf{Temporal graph neural networks (TGNN).} TGNNs learn from event streams with attention mechanisms, memory states, or evolving weights. These architectures generalize static graph operations to continuous time, utilizing diverse mechanisms to update node states as the network evolves~\cite{trivedi2019dyrep, xu2020tgat, rossi2020tgn, pareja2020evolvegcn, Zhou2020, wu2020comprehensive, Zhang2020deep, Cong2023}. DyRep parameterizes a temporal point process with dynamic embeddings to model both long and short term communications \cite{trivedi2019dyrep}, while TGAT uses continuous-time attention with functional encodings to prioritize recent neighbors \cite{xu2020tgat}. To enable inductive generalization, TGN introduces node memories and message passing that update at each event \cite{rossi2020tgn}. It specifically employs a trainable memory module to store the evolving state of each node, combined with a graph embedding module to generate temporal node representations. Given its flexible and modular design, TGN has become an established baseline in temporal graph learning and has inspired future architectures. Following this approach, GraphMixer simplifies the existing architectures entirely to reduce latency \cite{Cong2023}. Instead of TGN's attention mechanism, GraphMixer employs lightweight MLP-Mixer layers to efficiently encode temporal links while keeping computational costs low. With its simple and efficient architecture, GraphMixer achieves comparative state-of-the-art performance with faster convergence, and hence serves as a strong baseline for temporal link prediction. On the other hand, EvolveGCN takes a different approach by updating GCN weights through recurrent updates to track topological drift \cite{pareja2020evolvegcn}.  Despite their impressive empirical performance, TGNNs face several fundamental challenges. Most architectures require extensive hyperparameter tuning and are sensitive to initialization, making them difficult to deploy reliably across different domains \cite{wu2020comprehensive,Zhang2020deep}. The memory requirements often scale poorly with network size, limiting applicability to truly large-scale systems \cite{Zhou2020}. Additionally, most TGNNs focus on classification-style prediction tasks and do not naturally support sequential forecasting of interaction sequences. STEP addresses these limitations by reformulating the problem as a sequential forecasting task and leveraging discrete, probabilistic motif features to ensure scalability without the heavy memory overhead or training cost of neural network architectures. 

\noindent \textbf{Walk- and motif-based representations. } Structural representations capture the higher order interaction patterns to enable powerful inductive generalization. Wang et al. introduced Causal Anonymous Walks (CAW) to encode ordered paths while anonymizing node identities, providing compact inductive features for continuous-time graphs \cite{wang2021caw}. Temporal motifs summarize recurring time respecting patterns and related approaches show that motif counts can boost downstream performance with modest overhead \cite{zhao2022motif, chen2023tempme}. Specifically, TempME enhances model transparency by employing an information bottleneck framework to identify the most pivotal temporal motifs that drive a specific prediction~\cite{chen2023tempme}. Built on top of temporal graph neural networks as a modular add-on, TempME improves the performance of strong architectures such as TGAT, TGN and GraphMixer. With its motif-based prediction framework, it achieves state-of-the-art performance in temporal link prediction, and serves as our primary baseline in the experimental evaluation. 
Additional work has proposed motif mining as a signal for subgraph dynamics or anomaly detection \cite{Paranjape2017}. However, few models treat motifs as first-class elements in the event generation process, missing the opportunity to leverage them for efficient, structured prediction. STEP addresses these limitations by utilizing motif transition probabilities to capture latent structural information and support a generative framework in continuous time domain.

\section{Preliminaries}

In this work, we consider continuous-time temporal networks \cite{xiong2026survey}, defined as follows.
\begin{definition}[Temporal Graph] \label{def:temporal_graph}
A \emph{temporal graph} $G = (V, E)$ consists of a node set $V$ and a time-ordered event list $E$, where $\tau_{\max}$ denotes the timestamp of the last event. An \emph{event} $e_i = (u_i, v_i, t_i) \in E$ is a timestamped interaction between nodes $u_i$ and $v_i$ at time $t_i$. An \emph{edge} $(u,v)$ is a static connection between nodes $u$ and $v$; multiple events may occur on the same edge at different times. We designate $E'$ to the set of existing edges between the nodes in $G$.
\end{definition}

\begin{figure}[!t]
\includegraphics[width=8cm]{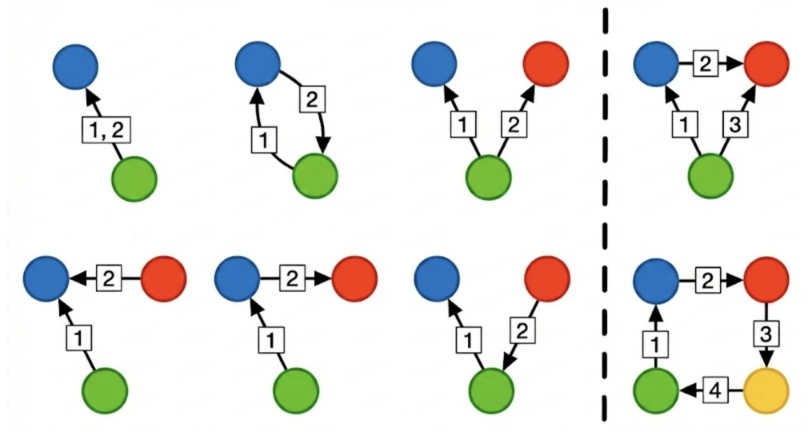}
\caption{\textbf{Examples of motif types in temporal graphs.} Event labels represent the temporal order of interactions and nodes are colored distinctly within each motif to indicate uniquness. The six motifs on the left forms the set of 2-event motifs, $\mathcal{M}_2$. The rightmost motifs illustrate an example 3-event and 4-event motif types.}
\label{img:motifs}
\end{figure}

\noindent In our model, we use the concept of Motif Transition Model (MTM) introduced by Liu and Sariyuce~\cite{liu2023mtm}, and we model event arrivals as a Poisson process. We adopt the temporal motif definition from \cite{liu2023mtm} which ensures that consecutive events are structurally connected to each other:
\begin{definition}[Temporal Motif] \label{def:motif}
Given a temporal graph $G = (V, E)$, an $\ell$-event temporal motif ($\ell \ge 2$), denoted by $m = (V_m, E_m)$, is a temporal subgraph of $G$ such that:
\begin{itemize}
    \item $V_m \subseteq V$, $E_m \subseteq E$, and $|E_m| = \ell$,
    \item $m$ is a weakly-connected subgraph, thus $2 \le |V_m| \le \ell + 1$,
    \item Each event is connected to at least one preceding event: $\{u_{j+1}, v_{j+1}\} \cap \{u_1, v_1, \ldots, u_j, v_j\} \neq \emptyset$ for any $j + 1 \le \ell$ (i.e., the motif remains a connected subgraph at every timestamp).
\end{itemize}
We use $\mathrm{type}(m)$ to express the type of motif $m$, which defines its unique structural pattern without reference to any node or timestamp. For a given $\ell$, we denote the set of all motif types of size $\ell$ by $\mathcal{M}_{\ell}$, or just $\mathcal{M}$ when $\ell$ is obvious.
The number of temporal motif types increases for larger values of $\ell$, e.g., there are 6 types motifs for $\ell=2$ ($|\mathcal{M}_2|=6$) and $|\mathcal{M}_3|=60$~\cite{liu2023mtm}.
Figure~\ref{img:motifs} illustrates a few examples of motifs.
\end{definition}

\noindent Next, we remind the concept of motif transition from \cite{liu2023mtm} that captures the evolution of a motif to another motif.
\begin{definition}[Motif Transition] \label{def:motif_transition}
In a temporal graph $G$, a motif $m = (V_m, E_m)$ transitions to a motif $m' = (V'_m, E'_m)$ if there exists a new event $e' = (u', v', t')$ such that:
\begin{itemize}
    \item $V'_m = V_m \cup \{u', v'\}$ and $E'_m = E_m \cup \{e'\}$,
    \item $\{u', v'\} \cap V_m \neq \emptyset$, i.e., the new event is adjacent to $m$,
    \item $t' > t_m$, where $t_m$ is the timestamp of the last event in $m$,
    \item $m$ does not transition to another motif before $e'$ arrives, i.e., there does not exist an event $e^* = (u^*, v^*, t^*) \in G$ that satisfies all requirements above and $t^* < t'$.
\end{itemize}
\end{definition}

\noindent Relatedly, we define the following cardinalities:

\begin{definition}[Edge Repetitions and Motif Transition Counts] \label{def:counts}
Let $C_e$ denote the count of event recurrences on edge $e$, and $C(x \to y)$ denote the number of transitions from motif type $x$ to motif type $y$.
\end{definition}

\noindent Next, we define the motif transition process to denote a series of transitions with respect to size and temporal limits:
\begin{definition}[Motif Transition Process] \label{def:motif_transition_process}
In a temporal graph $G$, a motif transition process is a sequence of motif transitions w.r.t a transition size limit, $\ell_{max}$ (number of events), and a transition time limit, $\Delta C$ (upper-bound on inter-event timings), such that:
\begin{itemize}
    \item The last motif in the sequence has $\ell_{max}$ events,
    \item The new event that creates the transition at each step is at most $\Delta C$ far from the last event,
    \item At each transition, there is no earlier event than the new event.
\end{itemize}
\end{definition}
 
\noindent Motif transition process captures well-defined micro evolutions in a temporal network, and serves as the fundamental building block that characterizes the rhythm of the network. Motif transition process also lets us distinguish all the events to two types:

\begin{definition}[Cold and Hot Events] \label{def:cold_hot}
Given the motif transition process definition above, an event $e = (u, v, t) \in G$ is classified as either:
\begin{itemize}
    \item A \emph{hot event} if there exists a motif $m$ such that $e$ causes $m$ to transition to $m'$, i.e., $e$ extends an existing motif instance.
    \item A \emph{cold event} if no such motif $m$ exists, i.e., $e$ initiates a new motif instance.
\end{itemize}
We define the cold event probability as $p_{\text{cold}} = \frac{|CE|}{|E|}$, where $CE$ denotes the set of all cold events in $G$.
\end{definition}

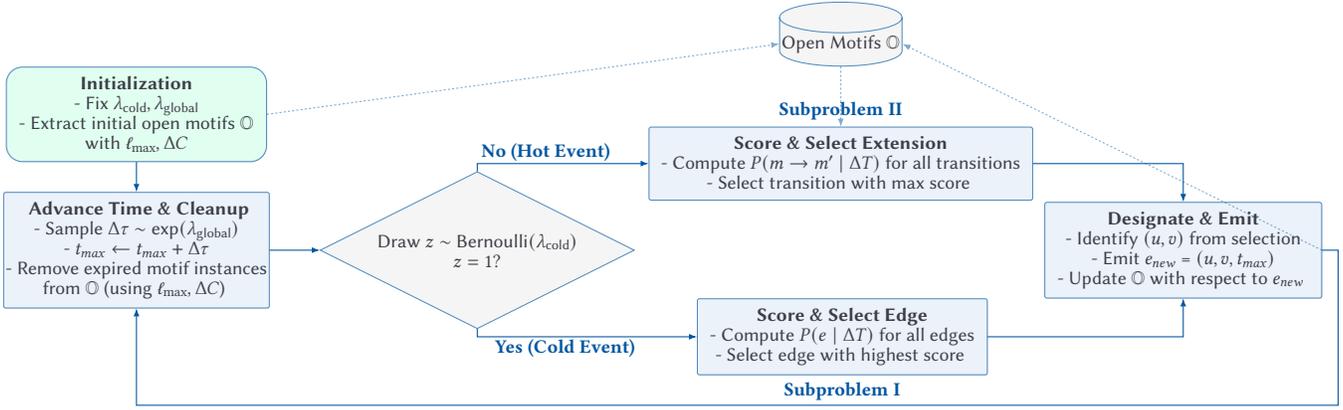
\begin{figure*}[htbp]
    \centering
    \resizebox{1.0\textwidth}{!}{
        \begin{tikzpicture}[
            node distance=1.6cm and 2.5cm,
            font=\fontsize{27}{28}\sffamily\color{stepdark},
            thick,
            process/.style={rectangle, draw=stepaccent!70, fill=stepblue, minimum width=3.5cm, minimum height=1.3cm, align=center, rounded corners=3pt, drop shadow={fill=black, opacity=0.1, shadow xshift=1pt, shadow yshift=-1pt}},
            inputblock/.style={rectangle, draw=stepaccent!70, fill=stepblue!40, minimum width=2.8cm, minimum height=1.2cm, align=center, rounded corners=6mm, drop shadow={fill=black, opacity=0.1}},
            decision/.style={diamond, draw=stepaccent!70, fill=stepgray, aspect=2, minimum width=3cm, align=center, inner sep=1pt},
            storage/.style={cylinder, shape border rotate=90, draw=stepaccent!70, fill=stepgray, aspect=0.25, minimum height=1.8cm, minimum width=1.6cm, align=center},
            arrow/.style={-{Latex[length=4mm, width=3mm]}, color=stepaccent, line width=1.5pt},
            dashed_arrow/.style={-{Latex[length=4mm, width=3mm]}, dashed, color=stepaccent, line width=1.5pt},
            line/.style={color=stepaccent, line width=1.5pt},
            container/.style={draw=stepaccent!50, dashed, fill=loopbg, inner sep=1cm, rounded corners=5pt}
        ]

            \node (time) [process, minimum width=4cm] {\\ \;\; \textbf{Advance Time \& Cleanup}\;\; \\- Sample $\Delta\tau \sim \text{exp}(\lambda_{\text{global}})$\\- $t_{max} \leftarrow t_{max} + \Delta\tau$\\- Remove expired motif instances\\ from $\mathbb{O}$ (using $\ell_{\max}, \Delta C$) \vspace{11pt}};
                        
            \node (init) [process, above=1.5cm of time, minimum width=4.5cm, fill=stepinit, rounded corners=0.8cm] {\\\textbf{Initialization}\\- Fix $\lambda_{\text{cold}}, \lambda_{\text{global}}$\\\;\;- Extract initial open motifs $\mathbb{O}$\;\;\\with $\ell_{\max}, \Delta C$\vspace{11pt}};
                        
            \draw [arrow] (init) -- (time);

            \node (decide) [decision, right=of time] {\;\;\;\;Draw $z \sim$ Bernoulli$(\lambda_{\text{cold}})$\;\;\;\;\\$z=1?$};

            \node (sp2_merged) [process, right=0.7cm of decide, yshift=4.3cm] {\\ \textbf{Score \& Select Extension}\\ \; - Compute $P(m \to m' \mid \Delta T)$ for all transitions \; \\- Select transition with max score \vspace{13pt}};
            
            \node [above=0.25cm of sp2_merged, font=\fontsize{27}{28}\bfseries\color{stepaccent}] {Subproblem II};

            \node (sp1_merged) [process, right=3.1cm of decide, yshift=-4.3cm] {\\\textbf{Score \& Select Edge}\\\;\;- Compute $P(e \mid \Delta T)$ for all edges\;\;\\- Select edge with highest score \vspace{11pt}};

            \node [below=0.2cm of sp1_merged, font=\fontsize{27}{28}\bfseries\color{stepaccent}] {Subproblem I};

            \node (storage) [storage, above=3cm of sp2_merged] {Open Motifs $\mathbb{O}$};

            \coordinate (midpoint) at ($(sp1_merged.east)!0.5!(sp2_merged.east)$);
            
            \node (emit) [process, right=1.7cm of midpoint, minimum height=1.5cm, minimum width=2.5cm] {\\ \textbf{Designate \& Emit}\\- Identify $(u,v)$ from selection\\- Emit $e_{new}$ = $(u, v, t_{max})$\\\;\;- Update $\mathbb{O}$ with respect to $e_{new}$\;\; \vspace{11pt}};

            \draw [arrow] (time) -- (decide);

            \draw [arrow] (decide) |- node[pos=0.7, above, font=\fontsize{27}{28}\bfseries, text=stepaccent] {No (Hot Event)} (sp2_merged);
            \draw [arrow] (decide) |- node[pos=0.7, below, font=\fontsize{27}{28}\bfseries, text=stepaccent] {Yes (Cold Event)} (sp1_merged);

            \draw [arrow] (sp1_merged) -| (emit);
            \draw [arrow] (sp2_merged) -| (emit);

            \draw [dashed_arrow, bend left=0, opacity=0.5] (init.east) to (storage.west);
            
            \draw [dashed_arrow, opacity=0.5] (storage) -- (sp2_merged);
            
            \draw [dashed_arrow, bend right=0, opacity=0.5] (emit.east) to (storage.east);

            \draw [line] (emit.east) -- ++(0.8,0) coordinate (far_right)
                |- ($(sp1_merged.south) - (0, 1.5)$) coordinate (bottom_path);
            
            \draw [arrow] (bottom_path) -| (time.south);

        \end{tikzpicture}
    }
    
\caption{The operational flow of the Stochastic Event Prediction (STEP) framework. After parameter initialization, the model enters a continuous prediction loop. In each iteration, time advances by $\Delta\tau$, and a Bernoulli decision determines whether to initiate a new motif (Subproblem I) or extend an existing motif (Subproblem II), updating the set of open motifs $\mathbb{O}$ accordingly.}
\vspace{-2ex}
\label{fig:step_flowchart}
\end{figure*}

\subsection{Poisson Process for Event Arrivals}

We assume that events arrive according to a Poisson process.
We define the intensity rates for edges and motif types based on observed inter-event time distributions.

\begin{definition}[Intensity Rate] \label{def:intensity_rate}
The \emph{intensity rate} $\lambda$ of a Poisson process represents the expected number of events per unit time. Equivalently, $1/\lambda$ is the expected waiting time between two consecutive events.
\end{definition}

Given an ordered sequence of timestamps $L = (t_1,\ldots,t_k)$, we calculate the intensity rate as
\[
f(L) = \left(\frac{1}{k-1}\sum_{i=2}^{k} (t_i - t_{i-1})\right)^{-1}.
\]
Intensity rate of a timestamp sequence is simply the inverse of the average inter-event time. We use $\lambda_{\text{global}} = f(\{t_i\})$ to denote the global intensity rate of a given temporal network; $\lambda_e = f(T_e)$ to show the intensity rate of an edge $e$, where $T_e$ is the ordered timestamp sequence of events occurring on $e$; and $\lambda_{s} = f(T_{s})\;$ to represent the intensity rate of a motif type $s=\mathrm{type}(m)$, where $T_{s}$ is constructed by collecting the timestamps of events that realize the evolution from a preceding motif instance to a motif of type $s$.

Under the Poisson model, the waiting time $\delta$ until the next recurrence on edge $e$ follows an exponential distribution with the following density: $p(\delta) = \lambda_e \exp\{-\lambda_e \delta\}$.
This follows from the memoryless property of the Poisson process: given that an event has occurred on edge $e$, the time until the next event on that edge is exponentially distributed with rate $\lambda_e$, independent of the history.

\section{STEP Framework} \label{sec:met}

We now present our framework, \emph{Stochastic Event Predictor} (STEP), for temporal link prediction in temporal graphs. STEP is governed by two core parameters: $\ell_{\max}$, the maximum number of events allowed in any motif instance, and $\Delta C$, the maximum allowed time between consecutive events within a motif. All motif construction and prediction decisions in STEP are defined according to these two parameters. The operational flow of STEP is illustrated in Figure~\ref{fig:step_flowchart}. The complete end-to-end computation process is detailed in Appendix \ref{ap:a}.

STEP models the evolution of a temporal graph by maintaining a dynamic set of \emph{open motifs} which are partial temporal motif instances that remain eligible for transition. Specifically, at any given time $\tau$, we track all open motifs that are eligible for extension.

\begin{definition}[Open Motif] \label{def:open_motifs}
A motif instance $m$ is considered \emph{open} at time $\tau$ if it satisfies two conditions:
\begin{itemize}
    \item The elapsed time since its most recent event is within a bounded temporal window: $\tau - \tau(m) \le \Delta C$,
    \item The motif has not yet reached the maximum allowed number of events: $|m| < \ell_{\max}$.
\end{itemize}
\end{definition}

\noindent We denote the set of all such motifs at time $\tau$ by $\mathbb{O}(\tau)$. These open motifs represent partial motif sequences that may still evolve as the network progresses.

To simulate future events, we follow a generative process that unfolds iteratively. At each step, the model:
\begin{enumerate}
    \item Samples a global inter-event delay $\Delta\tau \sim \mathrm{Exponential}(\lambda_{\text{global}})$ to determine when the next event occurs, then advances the current time to $\tau_{\max} \leftarrow \tau_{\max} + \Delta\tau$,
    \item Decides whether to start a new motif (cold event) or extend an existing one (hot event), based on a Bernoulli trial with success probability $p_{\text{cold}}$ (see Definition~\ref{def:cold_hot}).
    \item Depending on the Bernoulli trial, iterates over all candidate events and selects the one with highest unnormalized Bayes posterior.
\end{enumerate}

This framework reduces temporal link prediction to two subproblems: (1) starting a new motif instance, and (2) extending an existing open motif. We now describe each case in detail.

\subsection{Initiating a New Motif Instance}
\label{sec:initiation}

When a cold event is selected, the model identifies the most probable edge $e = (u, v)$ to initiate a new motif instance, conditioned on the waiting time $\Delta T = \tau_{\max}-\tau(e)$, where $\tau(e)$ denotes the last timestamp of the occurrence of edge $e$. This is done by computing the posterior score for each edge using Bayes' rule:
\[
P(e \mid \Delta T) \propto P(\Delta T \mid e) \cdot P(e),
\]
where $P(e)$ is the empirical prior computed as the fraction of events occurring on edge $e$ (see Definition \ref{def:counts}), and $P(\Delta T \mid e)$ is the likelihood of observing delay $\Delta T$ for edge $e$ under a Poisson process with rate $\lambda_e$ (see Definition~\ref{def:intensity_rate}). The likelihood is computed by integrating the exponential distribution over a small interval to approximate the probability mass around the event time, with $\epsilon$ fixed to 1 second which matches the timestamp granularity of typical datasets:
\begin{align*}
P(\Delta T \mid e) 
&= \int_{\Delta T - \epsilon}^{\Delta T + \epsilon} \lambda_e \exp\{-\lambda_e t\} \, dt \nonumber \\
&= \exp\{-\lambda_e (\Delta T - \epsilon)\} - \exp\{-\lambda_e (\Delta T + \epsilon)\}.
\end{align*}

For numerical stability, we use the unnormalized log-posterior score
\begin{equation}
\label{eq:cold_prob}
\begin{split}
\log P(e \mid \Delta T) 
&\propto \log \left( 
    \exp\{-\lambda_e (\Delta T - \epsilon)\} 
    - \exp\{-\lambda_e (\Delta T + \epsilon)\} 
   \right) \\
&\quad + \log \left( 
    \frac{C_e}{\sum_{e'} C_{e'}} 
   \right)
\end{split}
\end{equation}
and select the edge that maximizes this score. This edge is scheduled at time $\tau_{\max}$ and marks the start of a new motif instance. Note that we select cold events only from edges that have already appeared in the graph, rather than from all possible node pairs. This choice is consistent with our waiting time formulation: for edges that have not yet appeared in the graph, we have $\tau(e) = 0$, which would require a different treatment. We acknowledge that this design does not account for events on previously unseen node pairs and leave this extension to future work.

\subsection{Extending an Existing Open Motif}
If the model chooses to extend an existing motif, it evaluates all valid extensions of open motifs (see Definition~\ref{def:open_motifs}) at the current time $\tau_{\max}$. Each open motif $m \in \mathbb{O}(\tau_{\max})$ has a motif type $r = \mathrm{type}(m)$. The goal is to select a valid extension $m' = m \cup \{e\}$, where $e$ is a new event that transitions $m$ to $m'$ that satisfies Definition~\ref{def:motif_transition}. Additionally, we restrict these extensions to events between nodes that have already appeared in the motif transition process (see Definition~\ref{def:motif_transition_process}). This allows connections to form between known nodes that have not interacted before, but excludes events involving entirely unseen nodes (extending to unseen nodes would require a new mechanism that is more challenging to model, and hence it is left as a future work).

The model assigns a score to each possible extension $m \to m'$ by combining two terms: the likelihood of the waiting time $\Delta T = \tau_{\max}-\tau(m)$, where $\tau(m)$ designates the last timestamp of motif instance $m$, and the prior over motif type transitions. The likelihood that $\Delta T$ fits a Poisson process for the target motif type $s = \mathrm{type}(m')$ is given by:
\begin{align*}
P(\Delta T \mid m \to m') 
&= \int_{\Delta T - \epsilon}^{\Delta T + \epsilon} \lambda_{s} \exp\{-\lambda_{s} t\} \, dt \\
&= \exp\{-\lambda_{s} (\Delta T - \epsilon)\} - \exp\{-\lambda_{s} (\Delta T + \epsilon)\}.
\end{align*}
The prior probability of transitioning from motif type $r$ to $s$ is:
\begin{align*}
P(m \to m') = 
\frac{C(r \to s)}
{\sum_{t} C(r \to t)}
\end{align*}
where $C(x \to y)$ is the number of observed transitions between motif types in the training data (see Definition ~\ref{def:counts}). Combining both components, the (unnormalized) log-posterior score becomes:
\begin{equation}
\label{eq:hot_prob}
\begin{split}
\log P(m \to m' \mid \Delta T) \propto 
\log (
\exp\{-\lambda_{s} (\Delta T - \epsilon)\} 
 \\
- \exp\{-\lambda_{s} (\Delta T + \epsilon)\}  )
+ \log \left( 
\frac{C(r \to s)}
{\sum_{t} C(r \to t)} 
\right).
\end{split}
\end{equation}

\noindent The extension $m \to m'$ with the highest score is selected, and the corresponding event is added to the predicted event stream. The motif set $\mathbb{O}(\tau_{\max})$ is updated accordingly by removing the original motif instance $m$ and adding the extended instance $m'$, ensuring that only motifs relevant at time $\tau_{\max}$ remain open for potential future extensions. As in the first subproblem, $\epsilon$ is fixed to 1 second to match the timestamp granularity, which defines the integration bounds for computing the waiting time probability from the exponential density function.

\section{STEP as Feature Vectors} \label{sec:step_features}
While STEP is designed as a generative framework for sequential event prediction, its probabilistic modeling of motif transitions also yields broadly useful structural features. Since many existing temporal link prediction methods operate as binary classifiers that consume fixed-dimensional embeddings, we construct feature vectors that encode each event's structural and temporal context, enabling direct comparison and demonstrating that STEP can enhance these established approaches.

The key insight is that motif transitions capture how local interaction patterns evolve over time. Given that an event $e_i$ has occurred, we ask: from which open motif instances could this event have originated, and with what probability? Specifically, the STEP feature vector $\phi_{\text{STEP}}(e_i) \in \mathbb{R}^{|\mathcal{M}|}$ is indexed by motif type, where each entry reflects the posterior probability that $e_i$ resulted from extending an open motif with that entry's corresponding motif type. Events that complete common motif patterns with typical inter-event timing receive high probability mass on the corresponding motif types, while events representing unusual structural or temporal configurations yield more diffuse or low-magnitude vectors. This representation captures both the local connectivity context and the temporal plausibility of each event, providing complementary information to the existing TGNNs.

Full procedure (Algorithm~\ref{alg:step_embedding}) for generating STEP-based feature vectors is outlined in Appendix \ref{ap:b}. At each step, the algorithm maintains the pool of open motifs, identifies those that can be validly extended by the current event, computes the corresponding posterior probabilities based on temporal alignment, and updates the sparse feature matrix accordingly. This structured pipeline ensures that each event is encoded with respect to its local motif context and temporal positioning.

Given an event \( e_i \) occurring at time \( t_i \), we first identify the set of open motifs \( \mathbb{O}(t_i) \) that were valid and unexpired at that time (line~\ref{line:fv_expire}). From these, we extract the subset of motifs \( m \in \mathbb{O}(t_i) \) that can be legally extended to form a new motif instance \( m' \) by appending \( e_i \) (line~\ref{line:fv_check_valid}). This set represents all plausible motif continuations under the STEP framework at time \( t_i \).
For each such valid extension \( m \rightarrow m' \), we then compute the posterior probability that STEP would have chosen to make that extension given the delay \( \Delta T_m = t_i - \tau(m) \) (lines~\ref{line:delta_t} and \ref{line:fv_update_phi}).

We use this posterior to define a sparse feature vector \( \phi_{\mathrm{STEP}}(e_i) \in \mathbb{R}^{|\mathcal{M}|} \), indexed by motif type. Each nonzero entry corresponds to a motif \( m \) (with type $r$) that could be the preceding instance of a target motif \( m' \) (with type $s$), with its value normalized across all valid motif transitions:
\begin{equation} \label{eq:embed_coord}
\begin{aligned}
[\phi_{\mathrm{STEP}}(e_i)]_{\mathrm{index}(r)}
&= P\bigl(m\!\to\!m'\mid \Delta T_m\bigr) \\
&= \frac{
       P\bigl(\Delta T_m \mid m\!\to\!m'\bigr)\,
       P\bigl(m\!\to\!m'\bigr)
     }{
       \displaystyle\sum_{\substack{n \in \mathbb{O}(t_i) \\ n \cup \{e_i\} = n'}}
          P\bigl(\Delta T_n \mid n\!\to\!n'\bigr)\,
       P\bigl(n\!\to\!n'\bigr)
     }\,.
\end{aligned}
\end{equation}
Here, the temporal likelihood \( P(\Delta T_m \mid m \to m') \) and the motif transition prior \( P(m \to m') \) are defined in Equation~\eqref{eq:hot_prob}. For simplicity, we omit the edge-based cold event probability from the first subproblem (i.e., Equation~\eqref{eq:cold_prob}) in the feature construction, focusing only on motif extension dynamics. When the target motif $m'$ is reachable from multiple open motif instances of the same type, we aggregate their posterior probabilities into the single entry. The computation of these probabilities across all events results in a feature matrix \( \Phi \in \mathbb{R}^{|E| \times |\mathcal{M}|} \), where each row encodes the structure-aware representation of a specific event.

\section{Experiments} \label{sec:exps}
\textbf{Datasets.} We evaluate our approach on five benchmark temporal interaction datasets: CollegeMsg \cite{Panzarasa2009}, Email‐Eu \cite{Paranjape2017}, FBWall \cite{Viswanath2009}, SMS‐A \cite{Wu2010} and Wiki‐Talk \cite{Leskovec2014}. The CollegeMsg dataset comprises private messages exchanged among students over 193 days. Email-Eu contains 803 days of email communications within a European research institution. FBWall records user wall-post interactions on Facebook spanning 1,591 days. SMS-A consists of transactional SMS records covering 338 days. Wiki-Talk captures edit interactions on Wikipedia talk pages over 2,320 days. These datasets vary widely in scale from approximately 1K to over 1M unique nodes and in temporal volume from roughly 60K to 7.8M timestamped edges. Table~\ref{tab:dataset-stats} reports the precise node counts, temporal and static edge counts, and total duration for each dataset.

\begin{table}[!t]
  \centering
    \setlength{\tabcolsep}{1.5pt}
  \renewcommand{\arraystretch}{1.1}
  \begin{tabular}{l r r r r}
    \hline
    \textbf{Dataset} & \textbf{\# Nodes} & \textbf{\# Events} & \textbf{\# Edges} & \textbf{Timespan (days)} \\ \hline
    CollegeMsg       & 1,899             & 59,835                     & 20,296                   & 193                       \\
    Email-Eu         & 986               & 332,334                    & 24,929                   & 803                       \\
    SMS-A            & 44,430            & 548,182                    & 68,834                   & 338                       \\
    FBWall           & 46,799            & 859,050                    & 270,847                  & 1,591                     \\
    Wiki-Talk        & 1,140,149         & 7,833,140                  & 3,309,592                & 2,320                     \\ \hline
  \end{tabular}
  \caption{Summary statistics of the temporal graph datasets.}
  \label{tab:dataset-stats}
\end{table}

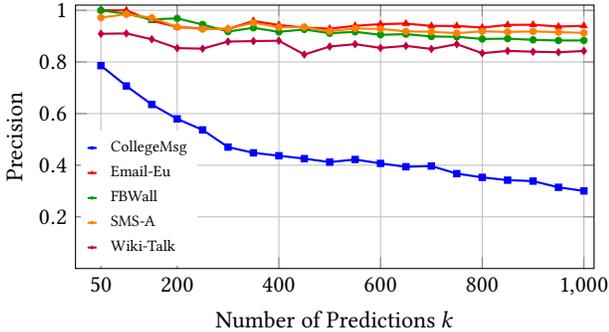
\begin{figure}[!t]
  \centering
  \begin{tikzpicture}
    \begin{axis}[
      width=\linewidth,
      height=0.6\linewidth,
      xlabel={Number of Predictions \(k\)},
      ylabel={Precision},
      xmin=0, xmax=1020,
      ymin=0, ymax=1.02,
      xtick={50,200,400,600,800,1000},
      ytick={0.2,0.4,0.6,0.8,1.0},
      grid=major,
      legend style={
        at={(0.01,0.01)},
        anchor=south west,
        font=\scriptsize,
        draw=none,
        cells={anchor=west},
        inner sep=2pt,
        column sep=3pt
      },
      legend image post style={scale=0.2},
    ]
    
      \addplot[
        color=blue,
        thick,
        mark=square*,
        mark size=1pt
      ] coordinates {
        (50,0.785714)  (100,0.706897) (150,0.635294) (200,0.579439)
        (250,0.536765) (300,0.470238) (350,0.447917) (400,0.436681)
        (450,0.425532) (500,0.411985) (550,0.422145) (600,0.407051)
        (650,0.394022) (700,0.396867) (750,0.367397) (800,0.352941)
        (850,0.342159) (900,0.338521) (950,0.314338) (1000,0.300518)
      };
      \addlegendentry{CollegeMsg}

      \addplot[
        color=red,
        thick,
        mark=triangle*,
        mark size=1pt
      ] coordinates {
        (50,1.000000)  (100,1.000000) (150,0.959596) (200,0.935484)
        (250,0.929577) (300,0.927778) (350,0.958525) (400,0.942857)
        (450,0.933086) (500,0.928814) (550,0.939873) (600,0.946429)
        (650,0.948509) (700,0.939547) (750,0.939394) (800,0.933185)
        (850,0.943434) (900,0.944123) (950,0.937153) (1000,0.939716)
      };
      \addlegendentry{Email-Eu}

      \addplot[
        color=green!60!black,
        thick,
        mark=*,
        mark size=1pt
      ] coordinates {
        (50,1.000000)  (100,0.986301) (150,0.964912) (200,0.968750)
        (250,0.945000) (300,0.917695) (350,0.932143) (400,0.916409)
        (450,0.925926) (500,0.911111) (550,0.916667) (600,0.905155)
        (650,0.907721) (700,0.898524) (750,0.897690) (800,0.888889)
        (850,0.890370) (900,0.885599) (950,0.883117) (1000,0.883273)
      };
      \addlegendentry{FBWall}

      \addplot[
        color=orange,
        thick,
        mark=otimes*,
        mark size=1pt
      ] coordinates {
        (50,0.971429)  (100,0.985075) (150,0.970588) (200,0.935252)
        (250,0.927374) (300,0.927536) (350,0.952000) (400,0.934641)
        (450,0.934985) (500,0.919355) (550,0.929648) (600,0.927739)
        (650,0.918455) (700,0.916832) (750,0.911488) (800,0.918478)
        (850,0.916256) (900,0.918138) (950,0.915452) (1000,0.912500)
      };
      \addlegendentry{SMS-A}

      \addplot[
        color=purple,
        thick,
        mark=diamond*,
        mark size=1pt
      ] coordinates {
        (50,0.909091)  (100,0.910256) (150,0.887931) (200,0.853503)
        (250,0.851485) (300,0.878906) (350,0.880702) (400,0.882175)
        (450,0.828729) (500,0.859694) (550,0.868889) (600,0.854251)
        (650,0.862205) (700,0.850365) (750,0.868825) (800,0.833856)
        (850,0.842814) (900,0.839609) (950,0.837584) (1000,0.842432)
      };
      \addlegendentry{Wiki-Talk}
    \end{axis}
  \end{tikzpicture}
  \caption{Precision of STEP sequence forecasting versus the number of predicted events \(k\). It remains high near \(k=100\) across all five datasets.}
  \label{fig:precision-vs-k}
\end{figure}

\noindent \textbf{Baselines. } We evaluate the impact of STEP feature vectors using state-of-the-art temporal graph neural networks such as TGN, GraphMixer, and TempME. All TGNN baselines are trained using an 80\% training, 5\% validation, and 15\% testing chronological split. We adopt parameter settings from \cite{chen2023tempme} as follows. For TGN, we use 2 attention heads and 2 layers. The number of MLPMixer layers for GraphMixer is set to 2 by default. The time encoding dimension is fixed at 100 and the output embedding dimension at 172. Training is performed using the Adam optimizer with a learning rate of \(10^{-3}\), a batch size of 512, and early stopping after 5 consecutive epochs without improvement in Average Precision, which measures the area under the precision-recall curve and is standard for evaluating temporal link prediction~\cite{Poursafaei2022}.

\noindent \textbf{STEP hyperparameters.} The STEP framework uses two hyperparameters defined in Section \ref{sec:met}: the maximum motif length $\ell_{\text{max}} = 3$, which balances expressiveness with computational efficiency, and the transition time limit $\Delta C$. For $\Delta C$, we compute the inter-event time between each pair of consecutive events sharing at least one node, then set $\Delta C$ to the maximum value observed across all such pairs in the dataset. This data-driven choice adapts to each dataset's interaction frequency without manual tuning. No other hyperparameters are required. 

\noindent \textbf{STEP + TGNN integration.} To assess whether STEP features provide complementary information to neural approaches, we concatenate the STEP feature vector with the link prediction probability produced by each TGNN baseline. This allows us to measure the marginal benefit of motif-based temporal features when combined with learned embeddings, without modifying the underlying TGNN architectures. We also report standalone STEP results for completeness. For the classification task, we train a 2-layer MLP.

All experiments are performed on a Linux operating system running on a machine with Intel(R) Xeon(R) Gold 6130 CPU processor at 2.10 GHz with 256 GB memory. We also use NVIDIA V100 16GB GPU to run TGNNs. We implemented the STEP framework in C++ for optimal computational performance and scalability. To integrate the TGNN-derived predictions, we employed inter-process communication between the C++ feature engineering pipeline and the Python-based TGNN predictors, which were implemented using PyTorch 2.3.0 with CUDA 12.1. We designate TLE (Time Limit Exceeded) for jobs that took longer than 3 days, and OOM (Out of Memory) for those that exceed the available memory. Our framework and pipeline is available at \href{https://github.com/ibahadiraltun/stochastic-event-prediction}{https://github.com/ibahadiraltun/stochastic-event-prediction}.

We present our experimental results as follows:
\begin{itemize}
    \item \textbf{Sequence prediction (Section~\ref{sec:sequence}):} We evaluate STEP's ability to forecast future event sequences, analyzing precision across varying prediction sizes and datasets.
    \item \textbf{Classification performance (Section~\ref{sec:classification}):} We assess the impact of STEP feature vectors when combined with TGN and GraphMixer, comparing against the motif-aware baseline TempME.
    \item \textbf{Runtime analysis:} In each section, we report end-to-end running times for both tasks, demonstrating STEP's computational efficiency.
\end{itemize}

\subsection{STEP sequence prediction} \label{sec:sequence}
Our primary contribution is formulating temporal link prediction as a sequential forecasting problem. We evaluate how accurately STEP can predict the next $k$ events given only the observed history.

\noindent \textbf{Evaluation Metrics. } In our sequence prediction experiments, STEP employs an iterative ranking procedure: at each step it selects the highest‐scoring candidate event, updates its open motifs to reflect this choice, and then proceeds to forecast the next event. After \(k\) iterations, we measure precision as the proportion of these \(k\) predicted events that statistically occur within the final \(p\%\) of the event stream. The choice of \(k\) and \(p\) can be adapted to different application requirements or network scales.

\begin{figure}[!t]
  \centering
  \begin{tikzpicture}
    \begin{axis}[
      width=\linewidth,
      height=0.6\linewidth,
      xlabel={Test Ratio (\%)},
      ylabel={Precision},
      xmin=0, xmax=21,
      ymin=0, ymax=1.02,
      xtick={1,5,10,15,20},
      ytick={0.4,0.6,0.8,1.0},
      grid=major,
      legend style={
        at={(0.99,0.01)},
        anchor=south east,
        font=\scriptsize,
        draw=none,
        cells={anchor=west},
        inner sep=2pt,
        column sep=3pt
      },
      legend image post style={scale=0.2},
    ]

      \addplot[
        color=blue,
        thick,
        mark=square*,
        mark size=1pt
      ] coordinates {
        (1,0.344828)  (2,0.431034)  (3,0.517241)  (4,0.534483)
        (5,0.620690)  (6,0.620690)  (7,0.620690)  (8,0.620690)
        (9,0.637931) (10,0.655172) (11,0.655172) (12,0.689655)
       (13,0.689655) (14,0.689655) (15,0.689655) (16,0.706897)
       (17,0.706897) (18,0.706897) (19,0.706897) (20,0.706897)
      };
      \addlegendentry{CollegeMsg}

      \addplot[
        color=red,
        thick,
        mark=triangle*,
        mark size=1pt
      ] coordinates {
        (1,0.822581)  (2,0.870968)  (3,0.919355)  (4,0.935484)
        (5,0.983871)  (6,0.983871)  (7,0.983871)  (8,0.983871)
        (9,0.983871) (10,0.983871) (11,1.000000) (12,1.000000)
       (13,1.000000) (14,1.000000) (15,1.000000) (16,1.000000)
       (17,1.000000) (18,1.000000) (19,1.000000) (20,1.000000)
      };
      \addlegendentry{Email-Eu}

      \addplot[
        color=green!60!black,
        thick,
        mark=*,
        mark size=1pt
      ] coordinates {
        (1,0.684932)  (2,0.849315)  (3,0.876712)  (4,0.876712)
        (5,0.931507)  (6,0.945205)  (7,0.945205)  (8,0.945205)
        (9,0.958904) (10,0.958904) (11,0.958904) (12,0.958904)
       (13,0.958904) (14,0.958904) (15,0.972603) (16,0.972603)
       (17,0.986301) (18,0.986301) (19,0.986301) (20,0.986301)
      };
      \addlegendentry{FBWall}

      \addplot[
        color=orange,
        thick,
        mark=otimes*,
        mark size=1pt
      ] coordinates {
        (1,0.761194)  (2,0.865672)  (3,0.955224)  (4,0.955224)
        (5,0.955224)  (6,0.955224)  (7,0.985075)  (8,0.985075)
        (9,0.985075) (10,0.985075) (11,0.985075) (12,0.985075)
       (13,0.985075) (14,0.985075) (15,0.985075) (16,0.985075)
       (17,0.985075) (18,0.985075) (19,0.985075) (20,0.985075)
      };
      \addlegendentry{SMS-A}

      \addplot[
        color=purple,
        thick,
        mark=diamond*,
        mark size=1pt
      ] coordinates {
        (1,0.717949)  (2,0.807692)  (3,0.833333)  (4,0.871795)
        (5,0.884615)  (6,0.897436)  (7,0.897436)  (8,0.897436)
        (9,0.897436) (10,0.897436) (11,0.897436) (12,0.897436)
       (13,0.897436) (14,0.897436) (15,0.897436) (16,0.897436)
       (17,0.897436) (18,0.897436) (19,0.897436) (20,0.897436)
      };
      \addlegendentry{Wiki-Talk}

    \end{axis}
  \end{tikzpicture}
  \caption{Precision of STEP sequence forecasting as a function of the test ratio, \(p\%\) and fixed \(k = 100\). Precision improves with larger held‐out fractions and stabilizes beyond 10\% in most datasets.}
  \label{fig:precision-vs-test-ratio-100}
\end{figure}
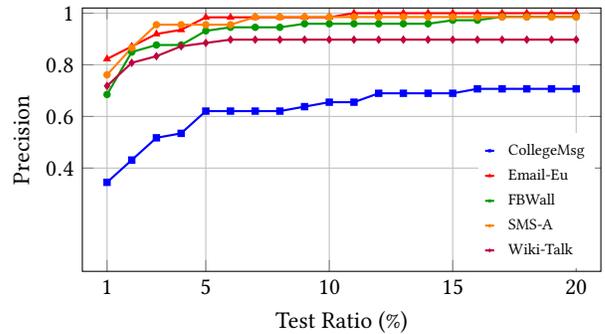

\noindent \textbf{Results.} We first fix the test ratio at $p = 20\%$ and vary the number of predictions $k$. Figure~\ref{fig:precision-vs-k} shows the change of the precision as $k$ increases from 50 to 1,000. It remains consistently high near $k=100$ for four of the five datasets. Moreover, larger graphs exhibit consistently high results as $k$ increases. Smaller relative drops are observed when predicting longer sequences. We then fix $k = 100$ and vary the test ratio $p$. Similarly, most datasets are stable after 10\% of the future event stream as shown in Figure~\ref{fig:precision-vs-test-ratio-100} and larger graphs achieve this stabilization faster. The smaller scale limits the availability of information needed to capture robust temporal patterns in the CollegeMsg dataset that experiences more pronounced performance drops, though precision still remains above 0.4 even at $k = 1000$.

\begin{table}[!t]
  \centering
  \setlength{\tabcolsep}{6pt}
  \renewcommand{\arraystretch}{1.1}
  \begin{tabular}{@{}l c c c@{}}
    \hline
    Dataset
      & RER
      & Node Entropy
      & Motif Trans. Entropy \\[4pt]
    \hline
    CollegeMsg
      & 24.59\%
      & 2.34
      & 2.62 \\
    Email-Eu
      & 90.08\%
      & 2.98
      & 2.35 \\
    FBWall
      & 48.02\%
      & 1.51
      & 2.36 \\
    SMS-A
      & 85.29\%
      & 0.62
      & 1.95 \\
    Wiki-Talk
      & 34.92\%
      & 0.72
      & 2.42 \\
    \hline
  \end{tabular}
  \caption{Repeated Event Ratio (RER), node entropy, and motif transition entropy statistics. RER measures the proportion of events that are repeated interactions and the entropy quantifies uncertainty at node and motif transition levels.}
  \label{tab:dataset-entropy-stats}
\end{table}

\noindent \textbf{Discussion.} To better understand the performance differences across datasets and quantify the uncertainty in the prediction process, Table~\ref{tab:dataset-entropy-stats} reports dataset-level coverage and entropy statistics. The repeated event ratio (RER) measures the fraction of future events (last 20\% percent) whose corresponding node pairs have been observed previously, measuring how much the new interactions happen between the node pairs that have not interacted before in the temporal network. A higher RER indicates more repetitive interaction patterns, which are easier to extrapolate. To quantify interaction uncertainty, we also report node entropy and motif transition entropy among the training events, both computed using Shannon entropy \cite{shannon1948}. Node entropy is defined by first computing, for each source node $u$, the entropy of its empirical target distribution,
\[H(u) = - \sum_{v \in \mathcal{N}(u)} p_{u,v} \log p_{u,v},\]
where $\mathcal{N}(u)$ denotes the set of distinct target nodes that $u$ has interacted with, $p_{u,v}$ denotes the relative frequency of interactions from $u$ to $v$, and the reported node entropy is obtained by averaging $H(u)$ across all nodes. We analogously define motif transition entropy as the entropy of the empirical distribution over transitions between motif types, capturing uncertainty in temporal motif extension.

Email-Eu and SMS-A display very high RERs of 90.08\% and 85.29\%, indicating that most future interactions happen on the node pairs that have interacted before. Combined with moderate to high motif transition entropy, this strong recurrence yields highly stable and near-perfect precision. FBWall similarly benefits from a moderate RER of 48.02\% together with low node entropy of 1.51, resulting in consistently high performance across prediction horizons.

In contrast, CollegeMsg exhibits substantially lower performance due to sparse edge recurrence and elevated structural uncertainty. Its RER is 24.59\%, meaning that the majority of future interactions involve previously unseen node pairs and require the model to generalize beyond historical edge patterns. This difficulty is compounded by a relatively high node entropy of 2.34 and the highest motif transition entropy of 2.62 among all datasets, reflecting heterogeneous participation and highly variable temporal motif evolution. Together, these factors lead to increased ambiguity in future event prediction and a faster degradation in precision as the number of predictions increases.

Wiki-Talk occupies an intermediate regime. Although its RER is 34.92\%, its interaction dynamics exhibit sufficient regularity to maintain reasonable precision. Additionally, lower node entropy assists in the performance of the hot event predictions. However, its motif transition entropy of 2.42 reflects complex temporal dynamics, which introduce additional uncertainty and limit peak performance compared to datasets with stronger edge-level recurrence.

\begin{table}[!t]
  \centering
  \setlength{\tabcolsep}{6pt}
  \renewcommand{\arraystretch}{1.1}
  \begin{tabular}{@{}l cc@{}}
    \hline
    Dataset      & Initialization time (s) & Time per prediction (ms) \\[4pt]
    \hline
    CollegeMsg   & 1.309                                  & 5.09                           \\
    Email-Eu     & 4.795                                  & 2.33                           \\
    FBWall       & 10.282                                 & 35.79                          \\
    SMS-A        & 31.013                                 & 10.05                          \\
    Wiki-Talk    & 42.760                                 & 784.06                         \\
    \hline
  \end{tabular}
  \caption{Runtime breakdown for STEP sequence forecasting.}
  \label{tab:efficiency-eval}
\end{table}

Overall, the interplay between edge recurrence and structural entropy fundamentally determines dataset predictability, with CollegeMsg representing the most challenging regime and Email-Eu and SMS-A the most favorable. Beyond dataset-level differences, performance also degrades as the number of predicted events increases, revealing an inherent trade-off between prediction horizon and forecast quality. However, as shown in Figure~\ref{fig:precision-vs-test-ratio-100}, precision stabilizes within approximately 10\% of the test stream across most datasets, with larger networks achieving stabilization faster, suggesting that STEP can reliably identify patterns from a modest fraction of future context.


Performance degrades as the number of predicted events increases, revealing an inherent trade-off between prediction size and forecast quality. However, as shown in Figure~\ref{fig:precision-vs-test-ratio-100}, precision stabilizes within approximately 10\% of the test stream across most datasets, with larger networks achieving stabilization faster. This observation suggests that STEP can reliably identify patterns from a modest fraction of future context.

\noindent \textbf{Runtime performance. }
We empirically evaluate the efficiency of STEP in terms of average inference time per predicted event. Table~\ref{tab:efficiency-eval} reports the mean wall-clock time required to generate a single sequence prediction across our five benchmark datasets. The motif initialization is one time pre-computation of motif transition counts $C(x \rightarrow y)$, edge occurrence counts $C_e$, intensity rates ($\lambda_{\text{global}}$, $\lambda_e$, $\lambda_s$), the initial set of open motifs $\mathcal{O}(\tau_{\max})$, and the cold event probability $p_{\text{cold}}$. These quantities are later used in iterative prediction process.

\subsection{STEP Classification} \label{sec:classification}
While STEP is designed primarily as a sequential forecasting framework, we also evaluate its effectiveness within the traditional binary classification paradigm to enable direct comparison with existing temporal graph neural network baselines. This allows us to assess whether STEP's motif-based probabilistic features provide complementary information that can enhance standard TGNN architectures without requiring modifications to their core designs.
\textbf{Evaluation metrics.} Following prior work on temporal link prediction \cite{chen2023tempme, Jin2024, Zheng2025}, we model the temporal link prediction problem as a binary classification task and evaluate model performance using Average Precision (AP).
This set-level precision metric evaluates whether predicted events appear in the future stream rather than whether their predicted ordering matches the ground truth. We adopt this formulation for consistency with prior work, which universally evaluates temporal link prediction through set-level metrics such as AP. Nonetheless, our setting is strictly harder than standard binary classification: the model must identify which specific edges will activate, rather than scoring predefined candidates. Evaluating order-sensitive metrics such as rank correlation remains an interesting direction for future work.

\begin{table}[!t]
  \centering
  \small
  \setlength{\tabcolsep}{1.5pt}
  \renewcommand{\arraystretch}{1.1}
  \begin{tabular}{@{}l| c | ccc | ccc@{}}
    \hline
      & 
      & \multicolumn{3}{c|}{\textbf{TGN}}
      & \multicolumn{3}{c}{\textbf{GraphMixer}} \\
    Dataset
      & STEP 
      & Base & \raisebox{0.57pt}{+}TempME & \raisebox{0.57pt}{+}STEP
      & Base & \raisebox{0.57pt}{+}TempME & \raisebox{0.57pt}{+}STEP \\
    \hline
    CollegeMsg
      & 83.76
      & 72.27
      & 68.71\textsuperscript{↓3.56}
      & \textbf{93.50}\textsuperscript{↑21.23}
      & 89.41
      & 86.75\textsuperscript{↓2.66}
      & \textbf{95.78}\textsuperscript{↑6.37} \\
    Email-Eu
      & 63.03
      & 87.24
      & 85.84\textsuperscript{↓1.39}
      & \textbf{90.31}\textsuperscript{↑3.07}
      & 68.80
      & 74.12\textsuperscript{↑5.32}
      & \textbf{80.90}\textsuperscript{↑12.1} \\
    FBWall
      & 75.51
      & 82.64
      & OOM
      & \textbf{92.22}\textsuperscript{↑9.58}
      & 88.11
      & OOM
      & \textbf{92.15}\textsuperscript{↑4.04} \\
    SMS-A
      & 79.39
      & 95.85
      & 96.62\textsuperscript{↑0.77}
      & \textbf{98.35}\textsuperscript{↑2.50}
      & 95.66
      & \textbf{97.98}\textsuperscript{↑2.32}
      & 97.44\textsuperscript{↑1.78} \\
    Wiki-Talk
      & 93.61
      & TLE
      & -
      & -
      & 97.06
      & OOM
      & \textbf{99.52}\textsuperscript{↑2.46} \\
    \hline
  \end{tabular}
  \caption{Average Precision (\%) scores on classification task. Dashes indicate unsupported experiments. The second column shows naive STEP results, obtained by a 2-layer MLP on STEP features. Best results are shown in bold.}
  \label{tab:classification-results}
\end{table}

\noindent \textbf{Results. } Table~\ref{tab:classification-results} presents the results.
Adding STEP feature vectors to temporal graph neural network models produces significant gains in average precision compared with the motif-aware baseline TempME. For the TGN, STEP raises average precision by 21.23\% on CollegeMsg and by 9.58\% on FBWall. Further improvements of 3.07\% and 2.50\% are recorded on Email-Eu and SMS-A, respectively. Under the GraphMixer model, STEP achieves increases of 12.10\% on Email-Eu and 6.37\% on CollegeMsg. In contrast, TempME fails to complete training on larger datasets such as FBWall and Wiki-Talk due to memory constraints, highlighting its limited scalability. This contrast suggests that STEP is not only more effective at leveraging temporal structure to increase predictive accuracy, but is also practical, and more robust on large-scale graphs.

STEP achieves strong standalone classification performance across benchmarks, with average precision scores of 83.76\% on CollegeMsg, 75.51\% on FBWall, 79.39\% on SMS-A, and 93.61\% on Wiki-Talk. While Email-Eu shows more modest performance at 63.03\%, it benefits substantially from combination with TGNN embeddings. These results demonstrate that motif-based probabilistic features are effective at capturing temporal patterns even without neural network integration with only two fixed hyperparameters and a lightweight probabilistic model.

\noindent \textbf{Runtime performance. } We further assess the computational overhead of each workflow by measuring end to end running time for feature extraction and link prediction across the five benchmark datasets. Table~\ref{tab:runtime-results} gives average wall clock times in seconds over 10 runs (5 runs for Wiki-Talk) for TGN and GraphMixer workflows. These results illustrate that STEP introduces lower overhead and consistently outperforms TempME in efficiency.

\begin{table}[!t]
\small
  \centering
  \setlength{\tabcolsep}{2pt}
  \renewcommand{\arraystretch}{1.1}
    \begin{tabular}{@{}l| r | rrr | rrr@{}}
    \hline
      & 
      & \multicolumn{3}{c|}{\textbf{TGN}}
      & \multicolumn{3}{c}{\textbf{GraphMixer}} \\
    Dataset
      & STEP
      & Base & \raisebox{0.57pt}{+}TempME & \raisebox{0.57pt}{+}STEP
      & Base & \raisebox{0.57pt}{+}TempME & \raisebox{0.57pt}{+}STEP \\
    \hline
    CollegeMsg
      & 96
      & 983    & 5,067   & \textbf{169}
      & 1,350  & 1,066   & \textbf{175} \\
    Email-Eu
      & 511
      & 5,545  & 29,663  & \textbf{1,080}
      & 7,749  & 8,252   & \textbf{1,036} \\
    FBWall
      & 1,454
      & 25,322 & OOM     & 5,113
      & 13,833 & OOM     & 4,032 \\
    SMS-A
      & 953
      & 10,796 & 82,275  & \textbf{3,053}
      & 12,413 & 11,604  & \textbf{4,260} \\
    Wiki-Talk
      & 12,215
      & -      & -       & -
      & 54,190 & OOM     & 31,728 \\
    \hline
  \end{tabular}
  \caption{End-to-end running time (in seconds) of TempME versus STEP in two workflows. Dashes indicate unsupported experiments. Best results are shown in bold.}
  \label{tab:runtime-results}
\end{table}

\noindent \textbf{Discussion. } The experimental results confirm that STEP consistently enhances downstream link prediction performance when combined with both TGN and GraphMixer architectures. A primary advantage of STEP is its lightweight design, which requires only two fixed hyperparameters and avoids the extensive memory overhead observed in TempME
By integrating local temporal motifs directly into the feature engineering process, STEP captures fine‐grained structural information that standard TGNNs may overlook. On the other hand, STEP's reliance on motif enumeration may limit its ability to capture very long‐range temporal dependencies, and the choice of expiration window could introduce sensitivity in domains with irregular interaction patterns, which is worth to consider in a future work. Larger values of $\ell_{\max}$ would enable richer motif patterns but increase computational cost and memory requirements, while extending $\Delta C$ could capture longer-range dependencies at the expense of increased candidate space during motif extension. Future extensions could explore adaptive motif selection and tighter integration with neural backends to mitigate these limitations.

\section{Conclusion}
 We introduced STEP, a framework that reformulates temporal link prediction as a sequential forecasting problem by combining Poisson-driven event modeling with Bayesian scoring over temporal motif transitions. STEP captures both the probabilistic dynamics of edge arrivals and the local structural context provided by temporal motifs, drawing on foundational work in dynamic networks \cite{Casteigts2012}, network motif theory \cite{Milo2002}, inductive graph learning \cite{Hamilton2017GraphSAGE,wu2020comprehensive}, and the motif transition model~\cite{liu2023mtm}. Integrating STEP with leading temporal graph neural networks yields consistent average precision improvements across five benchmarks, while its standalone generative mode achieves up to 0.99 precision in next $k$ sequential forecasting.\\

\noindent \textbf{Limitations.} STEP faces two primary limitations. First, motif extension depends on nodes already present in the base motif, limiting generalization to entirely new regions of the graph. In practice, repeated event ratios ranging from 24.59\% to 90.08\% across our datasets (Table~\ref{tab:dataset-stats}) indicate that a substantial fraction of future interactions occur between previously connected node pairs, mitigating this concern. Second, motif structures identified during training may become less representative as the network evolves, related to the issue of concept drift in streaming data~\cite{gama2014survey}. The same high recurrence rates suggest that interaction patterns tend to persist rather than vanish, allowing STEP to leverage learned structural patterns even as the network evolves. Extending STEP to handle unseen node pairs and adaptive motif vocabularies remains an important direction for future work.


\newpage
\bibliographystyle{ACM-Reference-Format}
\bibliography{references}


\begin{thebibliography}{43}


\ifx \showCODEN    \undefined \def \showCODEN     #1{\unskip}     \fi
\ifx \showISBNx    \undefined \def \showISBNx     #1{\unskip}     \fi
\ifx \showISBNxiii \undefined \def \showISBNxiii  #1{\unskip}     \fi
\ifx \showISSN     \undefined \def \showISSN      #1{\unskip}     \fi
\ifx \showLCCN     \undefined \def \showLCCN      #1{\unskip}     \fi
\ifx \shownote     \undefined \def \shownote      #1{#1}          \fi
\ifx \showarticletitle \undefined \def \showarticletitle #1{#1}   \fi
\ifx \showURL      \undefined \def \showURL       {\relax}        \fi
\providecommand\bibfield[2]{#2}
\providecommand\bibinfo[2]{#2}
\providecommand\natexlab[1]{#1}
\providecommand\showeprint[2][]{arXiv:#2}

\bibitem[Adamic and Adar(2003)]%
        {adamic2003}
\bibfield{author}{\bibinfo{person}{Lada~A. Adamic} {and} \bibinfo{person}{Eytan
  Adar}.} \bibinfo{year}{2003}\natexlab{}.
\newblock \showarticletitle{Friends and Neighbors on the Web}.
\newblock \bibinfo{journal}{\emph{Social Networks}} \bibinfo{volume}{25},
  \bibinfo{number}{3} (\bibinfo{year}{2003}), \bibinfo{pages}{211--230}.
\newblock
\href{https://doi.org/10.1016/S0378-8733(03)00009-1}{doi:\nolinkurl{10.1016/S0378-8733(03)00009-1}}


\bibitem[Asmussen and Glynn(2007)]%
        {asmussen2007}
\bibfield{author}{\bibinfo{person}{S{\o}ren Asmussen} {and}
  \bibinfo{person}{Peter~W. Glynn}.} \bibinfo{year}{2007}\natexlab{}.
\newblock \showarticletitle{Stochastic Simulation: Algorithms and Analysis}.
\newblock \bibinfo{journal}{\emph{Springer}} (\bibinfo{year}{2007}).
\newblock


\bibitem[Barab{\'a}si and Albert(1999)]%
        {barabasi1999}
\bibfield{author}{\bibinfo{person}{Albert-L{\'a}szl{\'o} Barab{\'a}si} {and}
  \bibinfo{person}{R{\'e}ka Albert}.} \bibinfo{year}{1999}\natexlab{}.
\newblock \showarticletitle{Emergence of Scaling in Random Networks}.
\newblock \bibinfo{journal}{\emph{Science}} \bibinfo{volume}{286},
  \bibinfo{number}{5439} (\bibinfo{year}{1999}), \bibinfo{pages}{509--512}.
\newblock
\href{https://doi.org/10.1126/science.286.5439.509}{doi:\nolinkurl{10.1126/science.286.5439.509}}


\bibitem[Casteigts et~al\mbox{.}(2012)]%
        {Casteigts2012}
\bibfield{author}{\bibinfo{person}{Arnaud Casteigts}, \bibinfo{person}{Paola
  Flocchini}, \bibinfo{person}{Walter Quattrociocchi}, {and}
  \bibinfo{person}{Nicola Santoro}.} \bibinfo{year}{2012}\natexlab{}.
\newblock \showarticletitle{Time-Varying Graphs and Dynamic Networks}.
\newblock \bibinfo{journal}{\emph{International Journal of Parallel, Emergent
  and Distributed Systems}} \bibinfo{volume}{27}, \bibinfo{number}{5}
  (\bibinfo{year}{2012}), \bibinfo{pages}{387--408}.
\newblock


\bibitem[Chen and Ying(2023)]%
        {chen2023tempme}
\bibfield{author}{\bibinfo{person}{Jialin Chen} {and} \bibinfo{person}{Rex
  Ying}.} \bibinfo{year}{2023}\natexlab{}.
\newblock \showarticletitle{TempME: Towards the Explainability of Temporal
  Graph Neural Networks via Motif Discovery}. In
  \bibinfo{booktitle}{\emph{NeurIPS 2023 Workshop on Temporal Graph Learning}}.
\newblock


\bibitem[Cong et~al\mbox{.}(2023)]%
        {Cong2023}
\bibfield{author}{\bibinfo{person}{Weilin Cong}, \bibinfo{person}{Si Zhang},
  \bibinfo{person}{Jian Kang}, \bibinfo{person}{Baichuan Yuan},
  \bibinfo{person}{Hao Wu}, \bibinfo{person}{Xin Zhou},
  \bibinfo{person}{Hanghang Tong}, {and} \bibinfo{person}{Mehrdad Mahdavi}.}
  \bibinfo{year}{2023}\natexlab{}.
\newblock \showarticletitle{Do We Really Need Complicated Model Architectures
  for Temporal Networks?}. In \bibinfo{booktitle}{\emph{International
  Conference on Learning Representations}}.
\newblock


\bibitem[Du et~al\mbox{.}(2016)]%
        {Du2016RMTPP}
\bibfield{author}{\bibinfo{person}{Nan Du}, \bibinfo{person}{Wei Dai},
  \bibinfo{person}{Rakshit Trivedi}, \bibinfo{person}{Urja Upadhyay},
  \bibinfo{person}{Manuel Gomez-Rodriguez}, {and} \bibinfo{person}{Le Song}.}
  \bibinfo{year}{2016}\natexlab{}.
\newblock \showarticletitle{Recurrent Marked Temporal Point Processes: A
  Recurrent Neural Network Approach for Predicting Events in Continuous Time}.
  In \bibinfo{booktitle}{\emph{Advances in Neural Information Processing
  Systems}}, Vol.~\bibinfo{volume}{29}. \bibinfo{pages}{1557--1565}.
\newblock


\bibitem[Gama et~al\mbox{.}(2014)]%
        {gama2014survey}
\bibfield{author}{\bibinfo{person}{Jo{\~a}o Gama}, \bibinfo{person}{Indr{\.e}
  {\v{Z}}liobait{\.e}}, \bibinfo{person}{Albert Bifet}, \bibinfo{person}{Mykola
  Pechenizkiy}, {and} \bibinfo{person}{Abdelhamid Bouchachia}.}
  \bibinfo{year}{2014}\natexlab{}.
\newblock \showarticletitle{A survey on concept drift adaptation}.
\newblock \bibinfo{journal}{\emph{ACM computing surveys (CSUR)}}
  \bibinfo{volume}{46}, \bibinfo{number}{4} (\bibinfo{year}{2014}),
  \bibinfo{pages}{1--37}.
\newblock


\bibitem[Grover and Leskovec(2016)]%
        {grover2016}
\bibfield{author}{\bibinfo{person}{Aditya Grover} {and} \bibinfo{person}{Jure
  Leskovec}.} \bibinfo{year}{2016}\natexlab{}.
\newblock \showarticletitle{node2vec: Scalable Feature Learning for Networks}.
  In \bibinfo{booktitle}{\emph{Proceedings of the 22nd ACM SIGKDD International
  Conference on Knowledge Discovery and Data Mining}}
  \emph{(\bibinfo{series}{KDD '16})}. \bibinfo{pages}{855--864}.
\newblock
\href{https://doi.org/10.1145/2939672.2939754}{doi:\nolinkurl{10.1145/2939672.2939754}}


\bibitem[Hamilton et~al\mbox{.}(2017)]%
        {Hamilton2017GraphSAGE}
\bibfield{author}{\bibinfo{person}{William~L. Hamilton}, \bibinfo{person}{Rex
  Ying}, {and} \bibinfo{person}{Jure Leskovec}.}
  \bibinfo{year}{2017}\natexlab{}.
\newblock \showarticletitle{Inductive Representation Learning on Large Graphs}.
  In \bibinfo{booktitle}{\emph{Proceedings of the 31st Conference on Neural
  Information Processing Systems}}. \bibinfo{pages}{1024--1034}.
\newblock


\bibitem[Hawkes(1971)]%
        {hawkes1971}
\bibfield{author}{\bibinfo{person}{Alan~G. Hawkes}.}
  \bibinfo{year}{1971}\natexlab{}.
\newblock \showarticletitle{Spectra of Some Self-Exciting and Mutually Exciting
  Point Processes}.
\newblock \bibinfo{journal}{\emph{Biometrika}} \bibinfo{volume}{58},
  \bibinfo{number}{1} (\bibinfo{year}{1971}), \bibinfo{pages}{83--90}.
\newblock
\href{https://doi.org/10.1093/biomet/58.1.83}{doi:\nolinkurl{10.1093/biomet/58.1.83}}


\bibitem[Huang et~al\mbox{.}(2023)]%
        {Huang2023}
\bibfield{author}{\bibinfo{person}{Shenyang Huang}, \bibinfo{person}{Farimah
  Poursafaei}, {and} \bibinfo{person}{Reihaneh Rabbany}.}
  \bibinfo{year}{2023}\natexlab{}.
\newblock \showarticletitle{Temporal Graph Benchmark for Machine Learning on
  Temporal Graphs}. In \bibinfo{booktitle}{\emph{NeurIPS 2023 Datasets and
  Benchmarks Track}}.
\newblock


\bibitem[Jin et~al\mbox{.}(2024)]%
        {Jin2024}
\bibfield{author}{\bibinfo{person}{Ming Jin}, \bibinfo{person}{Huan~Yee Koh},
  \bibinfo{person}{Qingsong Wen}, \bibinfo{person}{Daniele Zambon},
  \bibinfo{person}{Cesare Alippi}, \bibinfo{person}{Geoffrey~I Webb},
  \bibinfo{person}{Irwin King}, {and} \bibinfo{person}{Shirui Pan}.}
  \bibinfo{year}{2024}\natexlab{}.
\newblock \showarticletitle{A Survey on Graph Neural Networks for Time Series:
  Forecasting, Classification, Imputation, and Anomaly Detection}.
\newblock \bibinfo{journal}{\emph{IEEE Transactions on Pattern Analysis and
  Machine Intelligence}} (\bibinfo{year}{2024}).
\newblock
\href{https://doi.org/10.1109/TPAMI.2024.3443141}{doi:\nolinkurl{10.1109/TPAMI.2024.3443141}}


\bibitem[Katz(1953)]%
        {katz1953}
\bibfield{author}{\bibinfo{person}{Leo Katz}.} \bibinfo{year}{1953}\natexlab{}.
\newblock \showarticletitle{A New Status Index Derived from Sociometric
  Analysis}.
\newblock \bibinfo{journal}{\emph{Psychometrika}} \bibinfo{volume}{18},
  \bibinfo{number}{1} (\bibinfo{year}{1953}), \bibinfo{pages}{39--43}.
\newblock
\href{https://doi.org/10.1007/BF02289026}{doi:\nolinkurl{10.1007/BF02289026}}


\bibitem[Kumar et~al\mbox{.}(2019)]%
        {Kumar2019}
\bibfield{author}{\bibinfo{person}{Chaitanya Kumar}, \bibinfo{person}{Will
  Hamilton}, \bibinfo{person}{Jure Leskovec}, {and} \bibinfo{person}{Dan
  Jurafsky}.} \bibinfo{year}{2019}\natexlab{}.
\newblock \showarticletitle{Predicting Dynamic Embedding Trajectory in Temporal
  Interaction Networks}. In \bibinfo{booktitle}{\emph{Proceedings of the 25th
  ACM SIGKDD International Conference on Knowledge Discovery \& Data Mining}}.
  \bibinfo{pages}{1269--1278}.
\newblock


\bibitem[Lampert et~al\mbox{.}(2024)]%
        {Lampert2024}
\bibfield{author}{\bibinfo{person}{Moritz Lampert}, \bibinfo{person}{Shirui
  Pan}, {and} \bibinfo{person}{Ingo Scholtes}.}
  \bibinfo{year}{2024}\natexlab{}.
\newblock \showarticletitle{From Link Prediction to Forecasting: Addressing
  Challenges in Batch-based Temporal Graph Learning}.
\newblock \bibinfo{journal}{\emph{arXiv preprint arXiv:2406.04897}}
  (\bibinfo{year}{2024}).
\newblock
\href{https://doi.org/10.48550/arXiv.2406.04897}{doi:\nolinkurl{10.48550/arXiv.2406.04897}}


\bibitem[Leskovec and Krevl(2014)]%
        {Leskovec2014}
\bibfield{author}{\bibinfo{person}{Jure Leskovec} {and} \bibinfo{person}{Andrej
  Krevl}.} \bibinfo{year}{2014}\natexlab{}.
\newblock \bibinfo{title}{SNAP Datasets: Stanford Large Network Dataset
  Collection}.
\newblock
\newblock
\shownote{\url{http://snap.stanford.edu/data}}.


\bibitem[Liben-Nowell and Kleinberg(2003)]%
        {libenowell2003}
\bibfield{author}{\bibinfo{person}{David Liben-Nowell} {and}
  \bibinfo{person}{Jon Kleinberg}.} \bibinfo{year}{2003}\natexlab{}.
\newblock \showarticletitle{The Link-Prediction Problem for Social Networks}.
  In \bibinfo{booktitle}{\emph{Proceedings of the 12th International Conference
  on Information and Knowledge Management}} \emph{(\bibinfo{series}{CIKM
  '03})}. \bibinfo{pages}{556--559}.
\newblock
\href{https://doi.org/10.1145/956863.956972}{doi:\nolinkurl{10.1145/956863.956972}}


\bibitem[Liu and Sarıyüce(2023)]%
        {liu2023mtm}
\bibfield{author}{\bibinfo{person}{Penghang Liu} {and}
  \bibinfo{person}{Ahmet~Erdem Sarıyüce}.} \bibinfo{year}{2023}\natexlab{}.
\newblock \showarticletitle{Using Motif Transitions for Temporal Graph
  Generation}. In \bibinfo{booktitle}{\emph{Proceedings of the 29th ACM SIGKDD
  Conference on Knowledge Discovery and Data Mining}} (Long Beach, CA, USA)
  \emph{(\bibinfo{series}{KDD ’23})}. \bibinfo{publisher}{ACM},
  \bibinfo{pages}{1501--1511}.
\newblock
\href{https://doi.org/10.1145/3580305.3599540}{doi:\nolinkurl{10.1145/3580305.3599540}}


\bibitem[Mei and Eisner(2017)]%
        {mei2017neural}
\bibfield{author}{\bibinfo{person}{Hongyuan Mei} {and} \bibinfo{person}{Jason
  Eisner}.} \bibinfo{year}{2017}\natexlab{}.
\newblock \showarticletitle{Neural Hawkes Process: A New Model for Event
  Sequences}. In \bibinfo{booktitle}{\emph{Proceedings of the 34th
  International Conference on Machine Learning}} \emph{(\bibinfo{series}{ICML
  '17})}. \bibinfo{pages}{3488--3497}.
\newblock


\bibitem[Menon and Elkan(2011)]%
        {menon2011}
\bibfield{author}{\bibinfo{person}{Aditya~K. Menon} {and}
  \bibinfo{person}{Charles Elkan}.} \bibinfo{year}{2011}\natexlab{}.
\newblock \showarticletitle{Link Prediction via Matrix Factorization}. In
  \bibinfo{booktitle}{\emph{Proceedings of the 2011 European Conference on
  Machine Learning and Knowledge Discovery in Databases}}
  \emph{(\bibinfo{series}{ECML PKDD '11})}. \bibinfo{publisher}{Springer},
  \bibinfo{pages}{437--452}.
\newblock
\href{https://doi.org/10.1007/978-3-642-23780-5_28}{doi:\nolinkurl{10.1007/978-3-642-23780-5_28}}


\bibitem[Milo et~al\mbox{.}(2002)]%
        {Milo2002}
\bibfield{author}{\bibinfo{person}{Ron Milo}, \bibinfo{person}{Shalev
  Shen-Orr}, \bibinfo{person}{Shalev Itzkovitz}, \bibinfo{person}{Nadav
  Kashtan}, \bibinfo{person}{David Chklovskii}, {and} \bibinfo{person}{Uri
  Alon}.} \bibinfo{year}{2002}\natexlab{}.
\newblock \showarticletitle{Network Motifs: Simple Building Blocks of Complex
  Networks}.
\newblock \bibinfo{journal}{\emph{Science}} \bibinfo{volume}{298},
  \bibinfo{number}{5594} (\bibinfo{year}{2002}), \bibinfo{pages}{824--827}.
\newblock


\bibitem[Panzarasa et~al\mbox{.}(2009)]%
        {Panzarasa2009}
\bibfield{author}{\bibinfo{person}{Pietro Panzarasa}, \bibinfo{person}{Tore
  Opsahl}, {and} \bibinfo{person}{Kathleen~M. Carley}.}
  \bibinfo{year}{2009}\natexlab{}.
\newblock \showarticletitle{Patterns and dynamics of users' behavior and
  interaction: Network analysis of an online community}.
\newblock \bibinfo{journal}{\emph{Journal of the American Society for
  Information Science and Technology}} \bibinfo{volume}{60},
  \bibinfo{number}{5} (\bibinfo{year}{2009}), \bibinfo{pages}{911--932}.
\newblock


\bibitem[Paranjape et~al\mbox{.}(2017)]%
        {Paranjape2017}
\bibfield{author}{\bibinfo{person}{Ashwin Paranjape},
  \bibinfo{person}{Austin~R. Benson}, {and} \bibinfo{person}{Jure Leskovec}.}
  \bibinfo{year}{2017}\natexlab{}.
\newblock \showarticletitle{Motifs in Temporal Networks}. In
  \bibinfo{booktitle}{\emph{Proceedings of the Tenth ACM International
  Conference on Web Search and Data Mining}}. \bibinfo{pages}{601--610}.
\newblock


\bibitem[Pareja et~al\mbox{.}(2020)]%
        {pareja2020evolvegcn}
\bibfield{author}{\bibinfo{person}{Aldo Pareja}, \bibinfo{person}{Giacomo
  Domeniconi}, \bibinfo{person}{Jie Chen}, \bibinfo{person}{Tengfei Ma},
  \bibinfo{person}{Toyotaro Suzumura}, \bibinfo{person}{Hiroki Kanezashi},
  \bibinfo{person}{Tim Kaler}, \bibinfo{person}{Charles Leiserson}, {and}
  \bibinfo{person}{Tao Le}.} \bibinfo{year}{2020}\natexlab{}.
\newblock \showarticletitle{EvolveGCN: Evolving Graph Convolutional Networks
  for Dynamic Graphs}. In \bibinfo{booktitle}{\emph{Proceedings of the 34th
  AAAI Conference on Artificial Intelligence}} \emph{(\bibinfo{series}{AAAI
  '20})}. \bibinfo{pages}{5363--5370}.
\newblock


\bibitem[Perozzi et~al\mbox{.}(2014)]%
        {perozzi2014}
\bibfield{author}{\bibinfo{person}{Bryan Perozzi}, \bibinfo{person}{Rami
  Al-Rfou}, {and} \bibinfo{person}{Steven Skiena}.}
  \bibinfo{year}{2014}\natexlab{}.
\newblock \showarticletitle{DeepWalk: Online Learning of Social
  Representations}. In \bibinfo{booktitle}{\emph{Proceedings of the 20th ACM
  SIGKDD International Conference on Knowledge Discovery and Data Mining}}
  \emph{(\bibinfo{series}{KDD '14})}. \bibinfo{pages}{701--710}.
\newblock
\href{https://doi.org/10.1145/2623330.2623732}{doi:\nolinkurl{10.1145/2623330.2623732}}


\bibitem[Poursafaei and Rabbany(2022)]%
        {Poursafaei2022}
\bibfield{author}{\bibinfo{person}{Farimah Poursafaei} {and}
  \bibinfo{person}{Reihaneh Rabbany}.} \bibinfo{year}{2022}\natexlab{}.
\newblock \showarticletitle{Towards Better Evaluation for Dynamic Link
  Prediction}. In \bibinfo{booktitle}{\emph{NeurIPS 2022 Datasets and
  Benchmarks Track}}.
\newblock


\bibitem[Rossi et~al\mbox{.}(2020)]%
        {rossi2020tgn}
\bibfield{author}{\bibinfo{person}{Emanuele Rossi}, \bibinfo{person}{Benjamin
  Chamberlain}, \bibinfo{person}{Fabrizio Frasca}, \bibinfo{person}{Davide
  Eynard}, \bibinfo{person}{Federico Monti}, {and} \bibinfo{person}{Michael
  Bronstein}.} \bibinfo{year}{2020}\natexlab{}.
\newblock \showarticletitle{Temporal Graph Networks for Deep Learning on
  Dynamic Graphs}. In \bibinfo{booktitle}{\emph{Proceedings of the 37th
  International Conference on Machine Learning Workshop on Graph Representation
  Learning}}.
\newblock


\bibitem[Shannon(1948)]%
        {shannon1948}
\bibfield{author}{\bibinfo{person}{Claude~E. Shannon}.}
  \bibinfo{year}{1948}\natexlab{}.
\newblock \showarticletitle{A Mathematical Theory of Communication}.
\newblock \bibinfo{journal}{\emph{Bell System Technical Journal}}
  \bibinfo{volume}{27}, \bibinfo{number}{3} (\bibinfo{year}{1948}),
  \bibinfo{pages}{379--423}.
\newblock


\bibitem[Trivedi et~al\mbox{.}(2019)]%
        {trivedi2019dyrep}
\bibfield{author}{\bibinfo{person}{Rakshit Trivedi}, \bibinfo{person}{Mehrdad
  Farajtabar}, \bibinfo{person}{Prasenjeet Biswal}, {and}
  \bibinfo{person}{Hongyuan Zha}.} \bibinfo{year}{2019}\natexlab{}.
\newblock \showarticletitle{DyRep: Learning Representations over Dynamic
  Graphs}. In \bibinfo{booktitle}{\emph{Proceedings of the 7th International
  Conference on Learning Representations}} \emph{(\bibinfo{series}{ICLR '19})}.
\newblock


\bibitem[Viswanath et~al\mbox{.}(2009)]%
        {Viswanath2009}
\bibfield{author}{\bibinfo{person}{Bimal Viswanath}, \bibinfo{person}{Arjun
  Post}, \bibinfo{person}{Krishna~P. Gummadi}, {and} \bibinfo{person}{Alan
  Mislove}.} \bibinfo{year}{2009}\natexlab{}.
\newblock \showarticletitle{On the Evolution of User Interaction in Facebook}.
  In \bibinfo{booktitle}{\emph{WOSN ’09: Proceedings of the Second ACM
  Workshop on Online Social Networks}}. \bibinfo{pages}{37--42}.
\newblock


\bibitem[Wang et~al\mbox{.}(2021)]%
        {wang2021caw}
\bibfield{author}{\bibinfo{person}{Xinyi Wang}, \bibinfo{person}{Yizhou Kim},
  \bibinfo{person}{Yizhou Sun}, \bibinfo{person}{Xudong Li}, {and}
  \bibinfo{person}{Hanghang Qi}.} \bibinfo{year}{2021}\natexlab{}.
\newblock \showarticletitle{Inductive Representation Learning in Temporal
  Networks via Causal Anonymous Walks}. In
  \bibinfo{booktitle}{\emph{Proceedings of the 35th International Conference on
  Neural Information Processing Systems}} \emph{(\bibinfo{series}{NeurIPS
  '21})}.
\newblock


\bibitem[Wu et~al\mbox{.}(2010)]%
        {Wu2010}
\bibfield{author}{\bibinfo{person}{Yong Wu}, \bibinfo{person}{Chang Zhou},
  \bibinfo{person}{Jianxi Xiao}, \bibinfo{person}{Jürgen Kurths}, {and}
  \bibinfo{person}{Hans~Joachim Schellnhuber}.}
  \bibinfo{year}{2010}\natexlab{}.
\newblock \showarticletitle{Evidence for a Bimodal Distribution in Human
  Communication}.
\newblock \bibinfo{journal}{\emph{Proceedings of the National Academy of
  Sciences of the United States of America}} \bibinfo{volume}{107},
  \bibinfo{number}{44} (\bibinfo{year}{2010}), \bibinfo{pages}{18803--18808}.
\newblock
\href{https://doi.org/10.1073/pnas.1013140107}{doi:\nolinkurl{10.1073/pnas.1013140107}}


\bibitem[Wu et~al\mbox{.}(2020)]%
        {wu2020comprehensive}
\bibfield{author}{\bibinfo{person}{Zonghan Wu}, \bibinfo{person}{Shirui Pan},
  \bibinfo{person}{Fengwen Chen}, \bibinfo{person}{Guodong Long},
  \bibinfo{person}{Chengqi Zhang}, {and} \bibinfo{person}{Philip~S Yu}.}
  \bibinfo{year}{2020}\natexlab{}.
\newblock \showarticletitle{A comprehensive survey on graph neural networks}.
\newblock \bibinfo{journal}{\emph{IEEE transactions on neural networks and
  learning systems}} \bibinfo{volume}{32}, \bibinfo{number}{1}
  (\bibinfo{year}{2020}), \bibinfo{pages}{4--24}.
\newblock


\bibitem[Xiong et~al\mbox{.}(2026)]%
        {xiong2026survey}
\bibfield{author}{\bibinfo{person}{Jiafeng Xiong}, \bibinfo{person}{Ahmad
  Zareie}, {and} \bibinfo{person}{Rizos Sakellariou}.}
  \bibinfo{year}{2026}\natexlab{}.
\newblock \showarticletitle{A survey of link prediction in temporal networks}.
\newblock \bibinfo{journal}{\emph{SN Computer Science}} \bibinfo{volume}{7},
  \bibinfo{number}{1} (\bibinfo{year}{2026}), \bibinfo{pages}{100}.
\newblock


\bibitem[Xu et~al\mbox{.}(2020)]%
        {xu2020tgat}
\bibfield{author}{\bibinfo{person}{Da Xu}, \bibinfo{person}{Chuanwei Ruan},
  \bibinfo{person}{Serhat Korpeoglu}, \bibinfo{person}{Sushant Kumar}, {and}
  \bibinfo{person}{Kannan Achan}.} \bibinfo{year}{2020}\natexlab{}.
\newblock \showarticletitle{Inductive Representation Learning on Temporal
  Graphs}. In \bibinfo{booktitle}{\emph{Proceedings of the 8th International
  Conference on Learning Representations}} \emph{(\bibinfo{series}{ICLR '20})}.
\newblock


\bibitem[Zarezade et~al\mbox{.}(2017)]%
        {zarezade2017}
\bibfield{author}{\bibinfo{person}{Ali Zarezade}, \bibinfo{person}{Mahsa
  Rabbani}, \bibinfo{person}{Mehrdad Farajtabar}, \bibinfo{person}{Hamid~R.
  Rabiee}, {and} \bibinfo{person}{Le Song}.} \bibinfo{year}{2017}\natexlab{}.
\newblock \showarticletitle{Gyan: A Spatio-Temporal Point Process for
  Predicting Trajectories}. In \bibinfo{booktitle}{\emph{Proceedings of the
  31st AAAI Conference on Artificial Intelligence}}
  \emph{(\bibinfo{series}{AAAI '17})}. \bibinfo{pages}{2697--2703}.
\newblock


\bibitem[Zhang et~al\mbox{.}(2018)]%
        {Zhang2020deep}
\bibfield{author}{\bibinfo{person}{Muhan Zhang}, \bibinfo{person}{Zhichun Cui},
  \bibinfo{person}{Marion Neumann}, {and} \bibinfo{person}{Yixin Chen}.}
  \bibinfo{year}{2018}\natexlab{}.
\newblock \showarticletitle{An end-to-end deep learning architecture for graph
  classification}. In \bibinfo{booktitle}{\emph{Proceedings of the AAAI
  Conference on Artificial Intelligence}}, Vol.~\bibinfo{volume}{32}.
  \bibinfo{pages}{4438--4445}.
\newblock


\bibitem[Zhao et~al\mbox{.}(2022)]%
        {zhao2022motif}
\bibfield{author}{\bibinfo{person}{Tianqi Zhao}, \bibinfo{person}{Yuxuan He},
  \bibinfo{person}{Yu Wang}, {and} \bibinfo{person}{Le Song}.}
  \bibinfo{year}{2022}\natexlab{}.
\newblock \showarticletitle{Temporal Graph Motifs for Learning on
  Continuous-Time Dynamic Networks}. In \bibinfo{booktitle}{\emph{Proceedings
  of the 28th ACM SIGKDD International Conference on Knowledge Discovery and
  Data Mining}} \emph{(\bibinfo{series}{KDD '22})}.
  \bibinfo{pages}{2413--2423}.
\newblock
\href{https://doi.org/10.1145/3534678.3539390}{doi:\nolinkurl{10.1145/3534678.3539390}}


\bibitem[Zheng et~al\mbox{.}(2025)]%
        {Zheng2025}
\bibfield{author}{\bibinfo{person}{Yanping Zheng}, \bibinfo{person}{Liangliang
  Yi}, {and} \bibinfo{person}{Zhaoyang Wei}.} \bibinfo{year}{2025}\natexlab{}.
\newblock \showarticletitle{A survey of dynamic graph neural networks}.
\newblock \bibinfo{journal}{\emph{Frontiers of Computer Science}}
  \bibinfo{volume}{19} (\bibinfo{year}{2025}), \bibinfo{pages}{196323}.
\newblock
\href{https://doi.org/10.1007/s11704-024-3853-2}{doi:\nolinkurl{10.1007/s11704-024-3853-2}}


\bibitem[Zhou et~al\mbox{.}(2020b)]%
        {zhou2020taggen}
\bibfield{author}{\bibinfo{person}{Dawei Zhou}, \bibinfo{person}{Lecheng
  Zheng}, \bibinfo{person}{Jiawei Han}, {and} \bibinfo{person}{Jingrui He}.}
  \bibinfo{year}{2020}\natexlab{b}.
\newblock \showarticletitle{A data-driven graph generative model for temporal
  interaction networks}. In \bibinfo{booktitle}{\emph{Proceedings of the 26th
  ACM SIGKDD International Conference on Knowledge Discovery \& Data Mining}}.
  \bibinfo{pages}{401--411}.
\newblock


\bibitem[Zhou et~al\mbox{.}(2020a)]%
        {Zhou2020}
\bibfield{author}{\bibinfo{person}{Jie Zhou}, \bibinfo{person}{Ganqu Cui},
  \bibinfo{person}{Shengding Hu}, \bibinfo{person}{Zhengyan Zhang},
  \bibinfo{person}{Cheng Yang}, \bibinfo{person}{Zhiyuan Liu},
  \bibinfo{person}{Lifeng Wang}, \bibinfo{person}{Changcheng Li}, {and}
  \bibinfo{person}{Maosong Sun}.} \bibinfo{year}{2020}\natexlab{a}.
\newblock \showarticletitle{Graph neural networks: A review of methods and
  applications}.
\newblock \bibinfo{journal}{\emph{AI Open}}  \bibinfo{volume}{1}
  (\bibinfo{year}{2020}), \bibinfo{pages}{57--81}.
\newblock
\href{https://doi.org/10.1016/j.aiopen.2021.01.001}{doi:\nolinkurl{10.1016/j.aiopen.2021.01.001}}


\bibitem[Zuo et~al\mbox{.}(2020)]%
        {Zuo2020}
\bibfield{author}{\bibinfo{person}{Simiao Zuo}, \bibinfo{person}{Haoming
  Jiang}, \bibinfo{person}{Zichong Li}, \bibinfo{person}{Tuo Zhao}, {and}
  \bibinfo{person}{Hongyuan Zha}.} \bibinfo{year}{2020}\natexlab{}.
\newblock \showarticletitle{Transformer Hawkes Process}. In
  \bibinfo{booktitle}{\emph{Proceedings of the 37th International Conference on
  Machine Learning}}, Vol.~\bibinfo{volume}{119}.
  \bibinfo{pages}{11692--11702}.
\newblock


\end{thebibliography}
\clearpage

\appendix
\section{STEP Framework: End-to-end Computation} \label{ap:a}
Algorithm~\ref{alg:step} describes the full prediction process. At each iteration, the model samples an inter-event time \( \Delta \tau \sim \mathrm{Exponential}(\lambda_{\text{global}}) \) (line~\ref{line:sample_dt}), advances time \( t \) (line~\ref{line:advance_time}), and refreshes the set of active open motifs \( \mathbb{O} \) (line~\ref{line:remove_motifs}). A Bernoulli trial (line~\ref{line:draw_bernoulli}) with probability \( p_{\text{cold}} \) determines whether to initiate a new motif (cold event) or extend an existing one (hot event).

For cold events (line~\ref{line:cold_check}), the model evaluates \( P(e \mid \Delta T_e) \) using Eq.~\eqref{eq:cold_prob} for each candidate edge \( e \) (line~\ref{line:cold_prob}), selects the edge \( e^* \) with maximum probability, records the event (line~\ref{line:R_update}), and adds it to \( \mathbb{O} \) (line~\ref{line:add_cold_motif}).

For hot events (line~\ref{line:hot_start}), the model enumerates valid extensions \( m \rightarrow m' \) for each \( m \in \mathbb{O} \), computes \( \Delta T_m = t - \tau(m) \) (line~\ref{line:foreach_extension}), evaluates \( \log P(m \rightarrow m' \mid \Delta T_m) \) using Eq.~\eqref{eq:hot_prob} (line~\ref{line:compute_logprob}). It then, selects the highest-scoring transition, records the event (line~\ref{line:record_hot_event}), and updates \( \mathbb{O} \) (line~\ref{line:update_motifs}).

This iterative process continues until the model produces exactly \( n \) future events. The prediction counter is incremented after each event (line~\ref{line:increment_pred}), and the algorithm terminates when the desired prediction size is reached. The final output is the complete predicted event sequence (line~\ref{line:output}).

\begin{algorithm}[!t]
\caption{STEP: STochastic Event Predictor $(E, \mathbb{M}, n)$}\label{alg:step}
\begin{algorithmic}[1]
  \STATE \textbf{Input:} Observed events \(E\), motif instances \(\mathbb{M}\), prediction size\,\(n\) \label{line:input}
  \STATE \textbf{Output:} Predicted event sequence $R$
  \STATE \(\mathbb{O}\gets\{\,m\in\mathbb{M}: \tau_{\max}-\tau(m)\le\Delta C\}\) \hfill // extract open motifs \label{line:init_motifs}
  \STATE \(t\gets \tau_{\max}\), \(\mathit{count}\gets 0\) \label{line:init_time}
  \STATE \(R \gets \emptyset\)
  \WHILE{\(\mathit{count}<n\)} \label{line:while_start}
    \STATE \(\Delta\tau \sim\mathrm{Exponential}(\lambda_{\mathrm{global}})\) \hfill /* Sample waiting time and \label{line:sample_dt}
    \STATE \(t\gets t+\Delta\tau\) \hfill advance latest timestamp */ \label{line:advance_time}
    \STATE // remove expired motif instances
    \STATE \(\mathbb{O} \leftarrow \mathbb{O} \setminus \{m \in \mathbb{O} : t - \tau(m) > \Delta C \lor |m| \ge \ell_{\max}\}\) \label{line:remove_motifs}
    \STATE \(z\sim\mathrm{Bernoulli}(p_{\text{cold}})\) \hfill // cold or hot event decision \label{line:draw_bernoulli}
    \IF{\(z=1\)} \label{line:cold_check}
        \STATE // cold event scenario (subproblem I)
        \STATE \(\Delta T_e \leftarrow t - \tau(e),\; \forall e \in E'\) \hfill // edge-based waiting time
        \STATE // calculate posterior for each edge
        \STATE \(e^* \leftarrow \arg\max_{e \in E'} P(e \mid \Delta T_e)\) \text{  via Eq.~\eqref{eq:cold_prob}} \label{line:cold_prob}
        \STATE $R \leftarrow R \cup \{(e^*, t)\}$ \hfill // update predictions  \label{line:R_update}
        \STATE \(\mathbb{O} \leftarrow \mathbb{O} \cup \{(e^*, t)\}\) \hfill // update open motifs \label{line:add_cold_motif}
        \label{line:solve_first}
    \ELSE \label{line:hot_start}
        \STATE // hot event scenario (subproblem II)
        \STATE \(\Delta T_m \leftarrow t - \tau(m),\; \forall m \in \mathbb{O}\) // instance-based waiting time \label{line:foreach_extension}
        \STATE // calculate posterior for each valid motif transition
        \STATE \((m, m') \leftarrow \arg\max_{m \to m'}P(m \to m' \mid \Delta T_m)\) \text{   via Eq.~\eqref{eq:hot_prob}} \label{line:compute_logprob}
        \STATE // extract new event and update predictions
        \STATE $R \leftarrow R \cup \{(m' \setminus m, t)\}$ \label{line:record_hot_event}
        \STATE // remove previous instance and update open motifs
        \STATE \(\mathbb{O} \leftarrow (\mathbb{O} \setminus m) \cup m'\) \label{line:update_motifs}
    \ENDIF
    \STATE \(count \leftarrow count + 1\) \label{line:increment_pred}
  \ENDWHILE \label{line:while_end}
  \RETURN $R$ \label{line:output}
\end{algorithmic}
\end{algorithm}

\section{Feature Vectors} \label{ap:b}
While Section~\ref{sec:step_features} introduced the conceptual framework for STEP feature vectors, here we provide the complete algorithmic details. The key insight is to compute the posterior probability that each event $e_i$ resulted from extending a specific open motif type. This computation requires three steps. First, we maintain the set of open motifs $\mathbb{O}$. Second, we identify which motifs can be validly extended by $e_i$. Third, we compute the temporal likelihood of each extension. The resulting feature matrix $\Phi \in \mathbb{R}^{|E| \times |\mathcal{M}|}$ encodes both structural and temporal context. Each row represents an event and each column corresponds to a motif type. Algorithm~\ref{alg:step_embedding} details this process. It proceeds chronologically through the event sequence and updates the sparse feature matrix as each event is processed.

\begin{algorithm}[!t]
\caption{STEP Feature Vectors}\label{alg:step_embedding}
\begin{algorithmic}[1]
  \STATE \textbf{Input:} Observed events \(E\), motif types \(\mathcal{M}\)
  \STATE \textbf{Output:} Feature matrix \(\Phi \in \mathbb{R}^{|E| \times M}\)
  \STATE \(\Phi \leftarrow \mathbf{0}_{|E|\times M}\) \hfill // feature matrix \label{line:fv_init}
  \STATE \(\mathbb{O} \leftarrow \emptyset\) \hfill // open motifs \label{line:fv_open}
  \FOR{\(i = 1,\dots,|E|\)} \label{line:fv_loop}
    \STATE // remove expired motif instances
    \STATE \(\mathbb{O} \leftarrow \mathbb{O} \setminus \{m \in \mathbb{O} : t_i - \tau(m) > \Delta C \lor |m| \ge \ell_{\max}\}\) \label{line:fv_expire}
    \STATE // open motifs that can transition with $e_i$
    \STATE \(\mathbb{C} \gets \{m\in\mathbb{O}: m\cup\{e_i\} \text{ valid}\}\) \label{line:fv_check_valid}
    \FOR{each \(m\in\mathbb{C}\)} \label{line:fv_inner_loop}
      \STATE \(m' \gets m\cup\{e_i\}\) \hfill // form target motif
      \STATE \( \Delta T_m \gets t_i - \tau(m) \) \hfill // instance-based waiting time \label{line:delta_t}
      \STATE // calculate normalized posterior for the motif transition
      \STATE \(\Phi[i,\mathrm{index}(m')]\) $\gets$ \(P(m\to m' \mid \Delta T_m)\) \text{ via Eq.~\ref{eq:embed_coord}}
      \label{line:fv_update_phi}
      \STATE // remove previous instance and update open motifs
      \STATE \(\mathbb{O} \leftarrow (\mathbb{O} \setminus m) \cup m'\)  \label{line:fv_insert_new}
    \ENDFOR
  \ENDFOR
  \STATE \textbf{return} \(\Phi\) \label{line:fv_return}
\end{algorithmic}
\end{algorithm}

\end{document}